\documentclass[10pt,twocolumn,letterpaper]{article}

\usepackage{cvpr}              

\usepackage[dvipsnames]{xcolor}
\newcommand{\red}[1]{{\color{red}#1}}

\definecolor{cvprblue}{rgb}{0.21,0.49,0.74}
\usepackage[pagebackref,breaklinks,colorlinks,citecolor=cvprblue]{hyperref}
\newtheorem{definition}{Definition}
\newtheorem{lemma}{Lemma}
\usepackage{enumitem}
\usepackage{wrapfig}
\usepackage{multirow}
\usepackage[accsupp]{axessibility}

\usepackage[normalem]{ulem}

\newcommand{\minisection}[1]{\vspace{.06in}\noindent{\textbf{#1}}}

\newcommand{\tablestylesmaller}[2]{\setlength{\tabcolsep}{#1}\renewcommand{\arraystretch}{#2}\centering\footnotesize}

\def\paperID{5500} 
\def\confName{CVPR}
\def\confYear{2024}

\title{Classes Are Not Equal: An Empirical Study on Image Recognition Fairness}

\author{
	 Jiequan Cui$^{1}$ \hspace{4pt} Beier Zhu$^{1}$ \hspace{4pt} Xin Wen$^{2}$ \hspace{4pt} Xiaojuan Qi$^{2}$ \hspace{4pt} Bei Yu$^{3}$ \hspace{4pt} Hanwang Zhang$^{1}$ \vspace{.3em} \\
	 Nanyang Technological University$^{1}$ \hspace{5pt}
    The University of Hong Kong$^{2}$ \hspace{5pt} 
    The Chinese University of Hong Kong$^{3}$ \\
	\vspace{-10pt}
}

\begin{document}
\maketitle
\begin{abstract}
In this paper, we present an empirical study on image recognition unfairness, \emph{\ie}, extreme class accuracy disparity on balanced data like ImageNet. 
We demonstrate that classes are not equal and unfairness is prevalent for image classification models across various datasets, network architectures, and model capacities.
Moreover, several intriguing properties of fairness are identified. First, 
the unfairness lies in problematic representation rather than classifier bias distinguished from long-tailed recognition. 
Second, with the proposed concept of \textit{Model Prediction Bias},
we investigate the origins of problematic representation during training optimization.
Our findings reveal that models tend to exhibit greater prediction biases for classes that are more challenging to recognize.
It means that more other classes will be confused with harder classes. Then the False Positives (FPs) will dominate the learning in optimization, thus leading to their poor accuracy. 
Further, we conclude that data augmentation and representation learning algorithms improve overall performance by promoting fairness to some degree in image classification. Code is available at \url{https://github.com/dvlab-research/Parametric-Contrastive-Learning}.
\end{abstract}    
\section{Introduction}
\label{sec:intro}
In the past decade, significant advancements have been made in image recognition. Researchers have diligently explored a multitude of techniques to continually enhance the recognition capabilities of deep models, such as
data augmentations~\cite{zhang2017mixup, yun2019cutmix,cubuk2018autoaugment}, the evolution of model architectures~\cite{krizhevsky2012imagenet, simonyan2014very, he2016deep, dosovitskiy2020image}, and representation learning~\cite{oord2018representation, he2020momentum, 10130611, he2022masked}.
However, it's worth noting that most of these endeavors primarily focus on achieving state-of-the-art overall accuracy, often overlooking the goal of consistent performance across all interested classes.

\begin{figure}[t]
    \subfloat[CIFAR-100]{\includegraphics[width=0.16\textwidth]{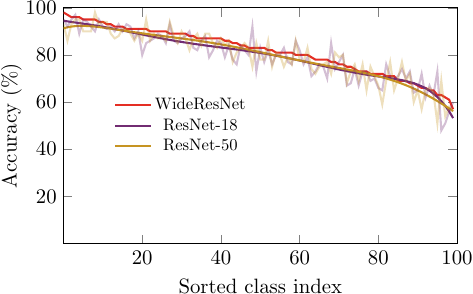}  \label{fig:fairness_cifar100}}
    \subfloat[ImageNet]{\includegraphics[width=0.16\textwidth]{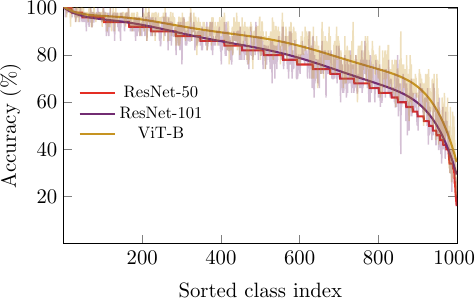}  \label{fig:fairness_imagenet}}
    \subfloat[WIT-400M]{\includegraphics[width=0.16\textwidth]{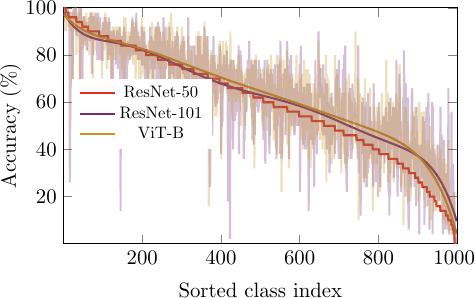}      \label{fig:fairness_clip}}
    \vspace{-0.1in}
    \caption{
        \textbf{The unfairness is prevalent in image classification across various datasets, network architectures, and model capacities.}
        We sort classes by the performance of WideResNet-34-10 on CIFAR-100 and ResNet-50 on ImageNet. For CLIP models, we sort classes by the zero-shot performance on ImageNet of the CLIP ResNet-50 model. Note that data rebalancing is considered in the collection of WIT-400M~\cite{radford2021learning}.
    }
\label{fig:fairness_cifar100_imagenet}
\end{figure}

 Recently, long-tailed recognition has garnered significant attention within the realms of computer vision and machine learning, due to its growing relevance in real-world applications. In practical scenarios, data often exhibits a long-tailed distribution, \ie, a few classes occupy plenty of data while most of the classes only have a few samples. Models trained on long-tailed data show extremely imbalanced performance. Especially, the accuracies of the low-frequency classes are pretty poor. 
 However, as shown in \Cref{fig:fairness_cifar100_imagenet}, we identify that a significant accuracy disparity can occur even on balanced datasets such as ImageNet~\cite{deng2009imagenet}. \textit{The best class achieves \textbf{100\%} top-1 accuracy while the worst class only achieves \textbf{16\%} top-1 accuracy with a ResNet-50 model on ImageNet.}
 This phenomenon suggests that factors beyond class frequency contribute to imbalanced performance. We hope that our findings will draw further attention from the research community to this pressing issue.

In this paper, we empirically investigate the image classification fairness issue, \ie, \textit{extreme accuracy disparity among classes}, on 8 balanced datasets including CIFAR and ImageNet. 
Our analysis also encompasses vision-language models, \ie, CLIP~\cite{radford2021learning} and the stable diffusion model~\cite{rombach2022high}. With a range of network architectures (CNNs and vision transformers), model capacities, and datasets, we identify this as a widespread issue in image classification models.

\noindent{\bf Representation or classifier bias?} 
Inspired by long-tailed recognition, we investigate the root of unfairness regarding classifier bias including $\ell_2$-norm of classifier weight and class separation angles. 
Challenging the conventional belief that larger separation angles correlate with higher accuracy, our findings reveal an intriguing contrast: \textit{classes with lower performance exhibit larger separation angles.}
Confirmed with $k$-nearest neighbors ($k$-NN) algorithm
and ETF classifier,
we conclude that the problematic representation serves as a crucial factor contributing to the unfairness.

\noindent{\bf Optimization challenges for problematic representation.}
We examine the underlying causes of problematic representation within the optimization process.
With the proposed concept of \textit{Model Prediction Bias}, 
our observations indicate a striking trend: \textit{The model displays higher prediction bias on classes that are more challenging to recognize}.
This trend exactly contradicts the principles of long-tailed recognition. 
It means that the harder the class, the more other classes will be confused with it, leading to False Positives (FPs) overwhelming True Positives (TPs) learning in the training optimization and thus their poor accuracy. 

\begin{figure}[t]
    \centering
    \includegraphics[width=0.46\textwidth]{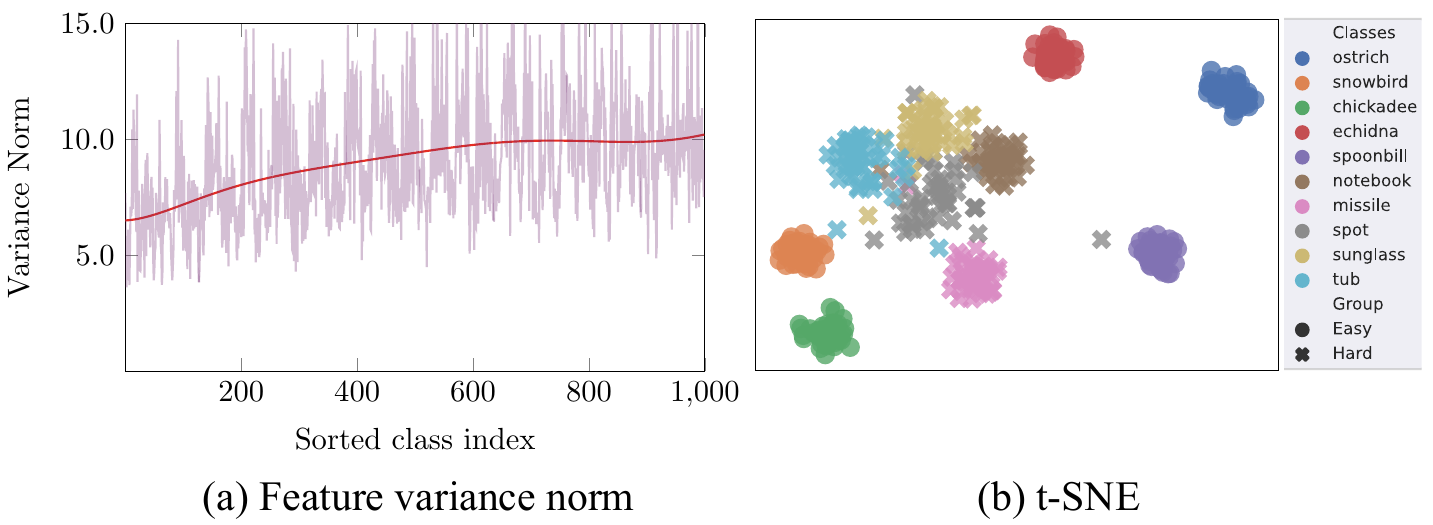}   
    \vspace{-0.1in}
    \caption{
     \textbf{Data diversity imbalance.}
        (a) Feature variance $\ell_2$-norm.
        (b) t-SNE visualization on ImageNet.
        We sort classes by the performance of ResNet-50 on ImageNet.
    }
\label{fig:l2_variance_r50_imagenet}
\end{figure}

Meanwhile, as shown in \Cref{fig:l2_variance_r50_imagenet}\red{(a)},
the increasing curve of ``the $\ell_2$-norm of the class mean feature variance'' implies that the feature distribution of hard classes is more diverse than that of easy classes.
The t-SNE visualization in \Cref{fig:l2_variance_r50_imagenet}\red{(b)} also indicates that the features of easy classes are more compact and separable than those of hard classes.
This phenomenon suggests that data associated with hard classes encompasses greater diversity,
covering more complex scenarios and resulting in overlaps with other classes in high probability.
Such overlaps can lead to class confusion in optimization. This is coherent with the analysis of model prediction bias. 

Finally, we study the techniques to improve fairness.
Data augmentations and representation learning algorithms promote overall performance, usually ignoring per-class accuracy.
With sound experiments, we observe that there are more performance gains in hard classes than in easy ones, leading to better fairness. A combination of data augmentations and representation learning tricks can boost fairness along with improved overall accuracy.

Our key contributions are as follows.
\begin{itemize}[leftmargin=15pt]
    \item We identify that the unfairness, \ie, extreme class accuracy disparity, is a general problem in image classification even on balanced data like ImageNet~\cite{deng2009imagenet}. 
    \item We study several intriguing properties of the unfairness phenomenon to understand it better. First, the extreme class accuracy disparity comes from problematic representation instead of biased classifiers. Second, with analysis of the model prediction bias, the harder the class is, the more other classes will be confused with it, thus leading to hard optimization and poor accuracy.
    \item We observe that data augmentations and representation learning algorithms achieve better fairness with improved overall accuracy while re-weighting sacrifices the accuracy of easy classes. 
\end{itemize}

\section{Related Work}

\minisection{Fairness}.
While unfairness is typically defined in terms of disparities related to sensitive attributes, like gender, race, disabilities, and sexual or political orientation, it is a growing concept.
In federated learning~\cite{DBLP:conf/iclr/LiSBS20}, fairness can be a measurement of the degree of uniformity in performance across federated client devices. For group distributionally robust optimization (DRO), spurious correlations are studied with group labels on the Waterbirds~\cite{sagawa2019distributionally}. In this paper, the unfairness is defined as the class accuracy disparities for image classification models.

\minisection{Imbalanced Learning}.
Re-weighting~\cite{cui2019class} and re-sampling~\cite{japkowicz2002class} are two classical methods to deal with data imbalance.
Since Cao et al.~\cite{cao2019learning} and Kang et al.~\cite{kang2019decoupling} observe that re-weighting and re-sampling can hurt the learned representation and thus degrade performance,
two-stage methods~\cite{zhong2021improving, zhang2021distribution} become popular, \ie, decoupling representation and classifier learning.
Wang et al.~\cite{wang2021longtailed} validate the effectiveness of model ensembling on long-tailed data.
Cui et al.~\cite{9774921} propose the residual mechanism for imbalanced learning.
After that, methods based on representation learning~\cite{cui2021parametric, 10130611} achieve new state-of-the-art results. 

\minisection{Data Augmentation and Representation Learning}.
Data augmentations~\cite{devries2017improved, cubuk2018autoaugment, cubuk2020randaugment, zhang2017mixup, yun2019cutmix} can significantly improve the generalization ability of models taking nearly no additional cost.
Single-image augmentations~\cite{devries2017improved, cubuk2018autoaugment, cubuk2020randaugment} do image-processing operations on each image independently. Cross-image augmentations~\cite{zhang2017mixup, yun2019cutmix} mix images and their labels simultaneously.
Representation learning is a fundamental task.
Contrastive loss~\cite{chen2020simple, he2020momentum} measures the similarities of sample pairs in feature space, and positive sample pairs are encouraged to have similar representations. 
Recently, masked modeling transferred from the neural language processing (NLP) community has made great progress in learning good representation for images, like~\cite{he2022masked,bao2021beit}, benefiting a wide spectrum downstream tasks.

Data augmentations and representation learning techniques both improve feature quality for recognition, thus promoting overall accuracy. However, per-class accuracy is usually ignored and we bridge the gap in this paper.

\begin{figure*}[t]
    \subfloat[Accuracy vs. Frequency]{ \includegraphics[width=0.23\textwidth]{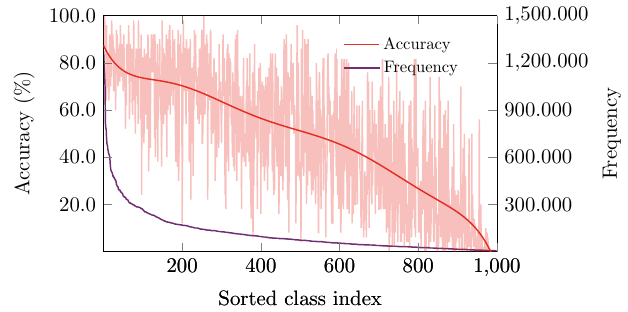}      \label{fig:imagenetlt_fre_acc}}
    \subfloat[$\ell_2$-norm of classifier]{ \includegraphics[width=0.19\textwidth]{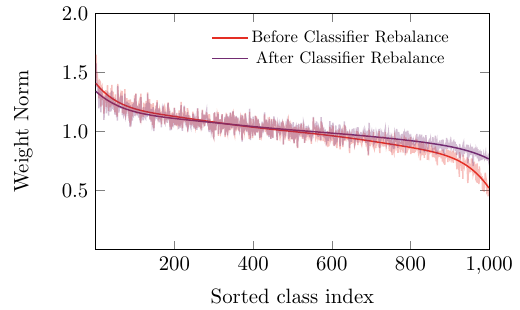}  \label{fig:weight_norm_lt}}
    \hspace{0.05in}
    \subfloat[``Many" of ImageNet-LT]{ \includegraphics[width=0.19\textwidth]{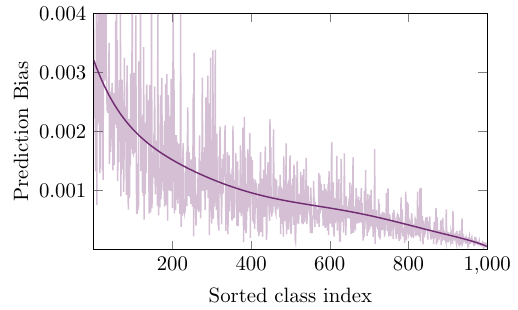}  \label{fig:imagenetlt_predictionbias_easy}}
    \subfloat[``Medium" of ImageNet-LT]{ \includegraphics[width=0.19\textwidth]{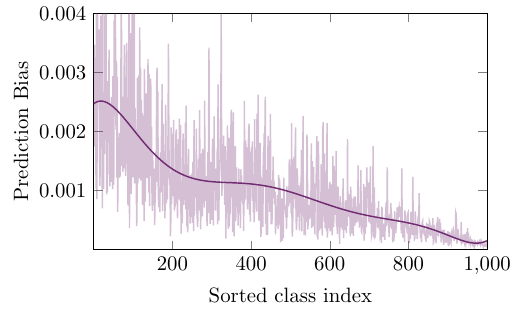}  \label{fig:imagenetlt_predictionbias_medium}}
    \subfloat[``Few" of ImageNet-LT]{ \includegraphics[width=0.19\textwidth]{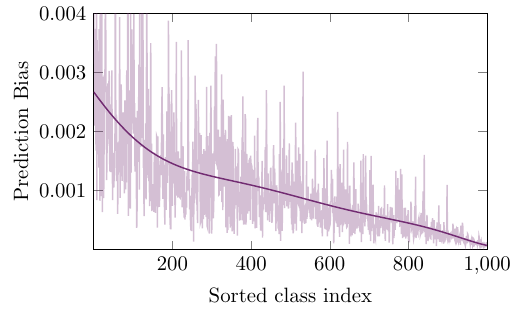}  \label{fig:imagenetlt_predictionbias_hard}}
    \vspace{-.5em}
    \caption{
        \textbf{Analysis on ImageNet-LT.}
        (a) Data distribution and per-class accuracy with ResNet-50 on ImageNet-LT;
        (b) The $\ell_2$-norm of classifier weights before/after classifier rebalancing;
        (c), (d), and (e) evaluate the prediction bias on ``Many", ``Medium", and ``Few" classes data separately.
        We sort classes by the number of samples in the classes on ImageNet-LT. 
    }
\label{fig:fairness_longtail}
\vspace{-0.1in}
\end{figure*}

\begin{figure*}[t]
    \subfloat[ImageNet-LT]{ \includegraphics[width=0.215\textwidth]{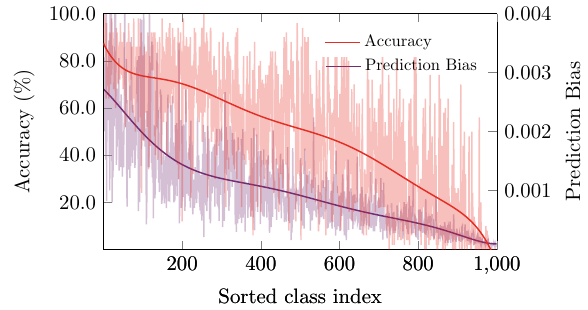}         \label{fig:imagenetlt_predition_bias}}
    \subfloat[ImageNet]{ \includegraphics[width=0.215\textwidth]{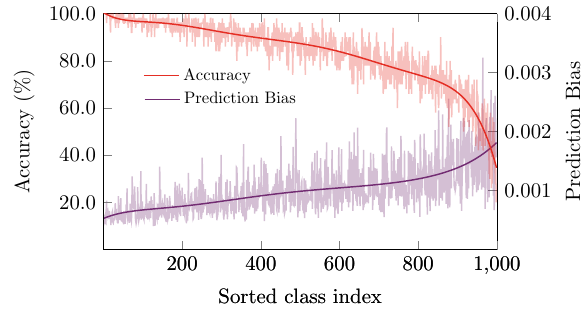}           \label{fig:imagenet_prediction_bias}}
    \subfloat[``Easy" of ImageNet]{ \includegraphics[width=0.19\textwidth]{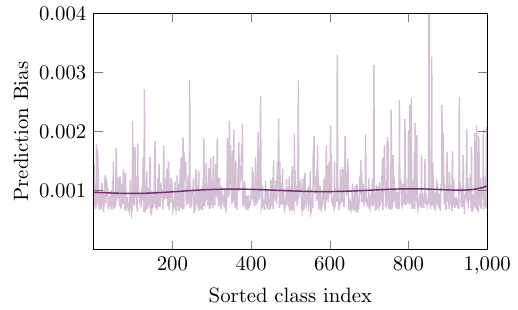}          \label{fig:imagenet_prediction_bias_easy}}
    \subfloat[``Meidum" of ImageNet]{ \includegraphics[width=0.19\textwidth]{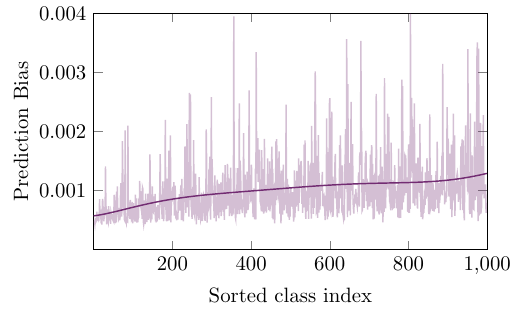}       \label{fig:imagenet_prediction_bias_medium}}
    \subfloat[``Hard" of ImageNet]{ \includegraphics[width=0.19\textwidth]{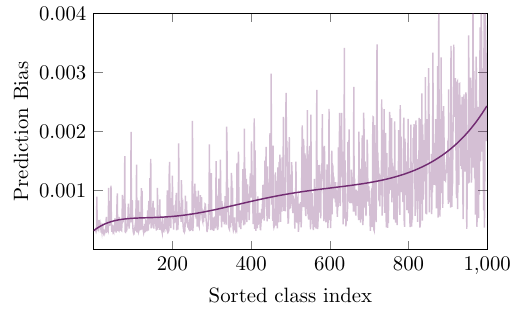}          \label{fig:imagenet_prediction_bias_hard}}
    \vspace{-.5em}
    \caption{
        \textbf{Model prediction bias.}
        (a) and (b) show the relationship between class accuracy and prediction bias. ImageNet-LT and ImageNet appear contrary conclusions. The details are discussed in \Cref{sec:prediction_bias}. 
        (c), (d), and (e) evaluate the prediction bias on ``Easy", ``Medium", and ``Hard" classes data separately.
        ResNet-50 is used for ImageNet-LT while ViT-B is adopted on ImageNet.
        Classes are sorted by their frequency on ImageNet-LT. On ImageNet, classes are sorted by the performance of ResNet-50.
    }
\label{fig:prediction_bias}
\vspace{-0.1in}
\end{figure*}

\begin{table}[t]
	\centering
	\caption{\textbf{Top-1 accuracy on ImageNet and CIFAR-100.} CLIP models trained on large-scale image-text pairs are also included.}
	\label{tab:fairness_imagenet_cifar100} 
    \vspace{-.5em}
    \tablestylesmaller{4.8pt}{1}
		\begin{tabular}{lcccccc}
			\toprule
			Method & Easy & Medium & Hard &All &Best  &Worst       \\
			\midrule
            \multicolumn{7}{c}{ImageNet} \\ 
            \midrule
			ResNet-50  &93.1 &81.1 &59.4  &77.8 &100.0 &16.0\\
            ResNet-101 &94.1 &82.6 &61.5  &79.4 &100.0 &16.0\\
            ViT-B      &95.4 &86.6 &68.9  &83.6 &100.0 &20.0\\
			\midrule
		  \multicolumn{7}{c}{CIFAR-100} \\
            \midrule
            ResNet-50        &89.8 &81.4 &68.5 &79.8 &98.0 &51.0\\
            WideResNet-34-10 &92.2 &83.2 &70.1 &81.7 &98.0 &57.0\\
            \midrule
		  \multicolumn{7}{c}{CLIP model on ImageNet} \\
            \midrule
            ResNet-50   &84.3 &61.8 &33.4 &59.8 &100.0 &0.0 \\
            ResNet-101  &83.6 &63.1 &40.3 &62.3 &100.0 &0.0 \\
			\bottomrule
		\end{tabular}
\end{table}

\section{Fairness Issue Widely Exists}
\label{sec:fairness_exists}
To showcase the widespread prevalence of fairness issues in image classification models, we conduct experiments on 8 datasets including CIFAR-100~\cite{krizhevsky2009learning} and ImageNet~\cite{russakovsky2015imagenet} with various network architectures including convolutional neural networks (CNNs) and vision transformers along with different model capacities. CLIP~\cite{radford2021learning} and the Stable Diffusion~\cite{rombach2022high} models trained on large-scale image-text pair data are also considered in our study. Please refer to the supplementary file for analysis of more datasets.

\minisection{Training on CIFAR-100}.
The CIFAR-100 dataset has 50,000 training images and 10,000 testing images with 100 classes. Each class has 500 training images and 100 testing images.
Following previous work~\cite{cui2019fast}, standard data augmentation including random crops with 4 pixels of padding and random horizontal flip is performed for data preprocessing in the course of training.

We use WideResNet34-10, and ResNet-50 as our backbones and follow the training schedule in~\cite{wu2020adversarial, cui2023decoupled}. We train the models 200 epochs with a cosine learning rate strategy and a batch size of 128 on 4 GPUs. The initial learning rate is set to 0.2. A weight decay of 5e-4 is adopted. We use the SGD optimizer with a momentum of 0.9 in training.

\minisection{Training on ImageNet}.
ImageNet~\cite{russakovsky2015imagenet} is one of the most challenging datasets for classification,
which consists of 1.2 million images for training and 50K images for validation over 1,000 classes.
It is worth noting that ImageNet is a carefully curated dataset. Data is balanced over all classes and each class has nearly 1000 training samples.

We use CNNs (ResNet-50 and ResNet-101) and vision transformers (ViT-B) as our backbones. For CNNs, we follow the training schedule from~\cite{yun2019cutmix}.
Models are trained in 300 epochs with a cosine learning rate strategy and a batch size of 256 on 8 GPUs. The initial learning rate and the weight decay are set to 0.1 and 1e-4 separately. The SGD optimizer with a momentum of 0.9 is used.

For training with ViT-B, we follow the training schedule in~\cite{he2022masked}, \ie, firstly pre-train it with masked modeling technique and then fine-tune it 100 epochs. Our implementation is built on their open-sourced code.

\minisection{Evaluation}.
In the long-tailed recognition field, ``Many", ``Medium", and ``Few" classes are defined for evaluation according to the number of samples in the classes. Similarly, to quantity the unfairness, we define ``Hard", ``Medium", and ``Easy" classes with a well-trained reference model.
ResNet50 and WideResNet34-10 are adopted as the reference models for ImageNet and CIFAR-100 separately.
The number of classes in ``Hard", ``Medium", and ``Easy" are evenly divided according to the performance of the reference model.
For CLIP~\cite{radford2021learning} models, we directly evaluate their pre-trained model on the ImageNet validation set and sort the classes with the performance of the ResNet-50 CLIP model. 

\minisection{Results Summary}.
We summarize the experimental results in \Cref{fig:fairness_cifar100_imagenet} and \Cref{tab:fairness_imagenet_cifar100}.
As plotted in \Cref{fig:fairness_cifar100_imagenet}, from easy to hard classes, the accuracy significantly decreases for models trained on CIFAR-100, ImageNet, and WIT-400M.
The phenomenon is consistent across various backbones including CNNs and vision transformers. Especially, the best class ``ostrich" achieves \textbf{100\%} top-1 accuracy while the worst class ``screen" only obtains \textbf{16\%} top-1 accuracy on ImageNet with ResNet-50, which demonstrates a large performance disparity on balanced data.

As shown in \Cref{tab:fairness_imagenet_cifar100}, ``Easy" classes can get much better performance than ``Hard" classes although training samples are evenly distributed among all the classes. For the CLIP ResNet-50 model, the average accuracy of ``Easy" classes is 50\% higher than that of ``Hard" classes.
The huge accuracy gap between the best (\textbf{100\%}) and the worst (\textbf{0\%}) classes indicates a surprising performance imbalance on balanced datasets. It is worth noting that data rebalancing strategies have already been applied in the collection of WIT-400M. Analysis of the other five datasets is included in the supplementary file.

With these experimental analyses, we conclude that the fairness issue in image classification is a general problem across various datasets, network architectures, and model capacities. Further, the similar trends on the same datasets with various models demonstrate that the unfairness highly depends on training data distribution. We expect that our study could raise more attention from the community to this issue and facilitate progress. 

\minisection{Discussion}.
When trained with long-tailed data, models tend to perform well on high-frequency classes but show significantly reduced accuracy on low-frequency ones. This pattern suggests to us that class frequency can be the primary driver of the extremely imbalanced performance. 
Although several works start to consider hard example learning,  most works~\cite{cao2019learning, kang2019decoupling, cui2019class, 9774921} in long-tailed recognition still focus on this aspect.
However, \textit{our research reveals that significant performance disparities (100\% accuracy on the best class while only 16\% on the worst class on ImageNet) also occur in balanced datasets. This indicates the presence of other significant factors that influence per-class accuracy, beyond just class frequency.}

\section{Analysis on Image Recognition Fairness}
\label{sec:fariness_imbalance}
As shown in \Cref{fig:fairness_cifar100_imagenet} and \Cref{fig:fairness_longtail} (a), extreme performance disparity happens both in long-tailed recognition and unfairness.
We delve into the properties of fairness in image classification by comparing it with long-tailed recognition.
The discussion on whether bias in representation or classifier leads to unfairness is presented in \Cref{sec:representation_classifier}.
Following that, we conduct an in-depth examination of the underlying optimization challenges for fairness in \Cref{sec:prediction_bias}.

\subsection{Representation or Classifier Bias?}
\label{sec:representation_classifier}
Inspired by long-tailed recognition, we diagnose the classifier bias for unfairness in terms of their weights $\ell_2$-norm and class separation angles. Then we conclude that the problematic representation results in unfairness.

\minisection{$\mathbf{\ell_2}$-norm of Classifier Weights}.
Previous works~\cite{cao2019learning, kang2019decoupling, zhong2021improving} study the classifier bias in long-tailed learning and observe $\ell_2$-norm of classifier weights matters. 
Kang et al.~\cite{kang2019decoupling} propose a two-stage pipeline: uniform data sampling for representation learning, and then class-balanced re-sampling for classifier learning with previously learned representation fixed.
With the method, we train a ResNet-50 model on ImageNet-LT~\cite{liu2019large}, which obeys the long-tailed distribution shown in \Cref{fig:imagenetlt_fre_acc}.

With the uniform sampling strategy for representation learning, the $\ell_2$-norm of the learned classifier weights positively correlates to class accuracy and class frequency as plotted in \Cref{fig:imagenetlt_fre_acc} and \Cref{fig:weight_norm_lt}. 
After classifier rebalancing using the re-sampling strategy, the norm of classifier weights becomes more uniform, which is visualized in \Cref{fig:weight_norm_lt}, and the accuracies of low-frequency classes are promoted.

\begin{figure}[tb!]
    \subfloat[Classifier $\ell_2$-norm on ImageNet]{ \includegraphics[width=0.22\textwidth]{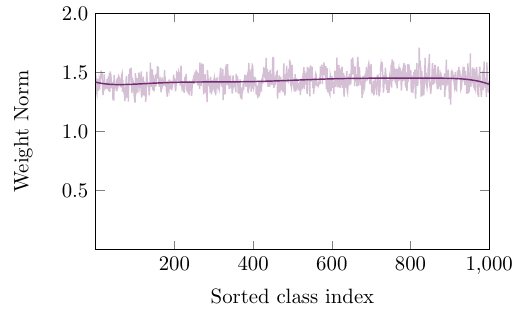}  \label{fig:imagenet_weightnorm}}
    \subfloat[CMA on ImageNet]{ \includegraphics[width=0.22\textwidth]{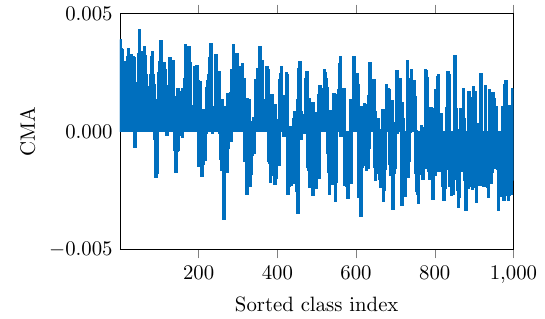}  \label{fig:imagenet_mean_margin}} \\
    \subfloat[$k$-NN on ImageNet]{ \includegraphics[width=0.22\textwidth]{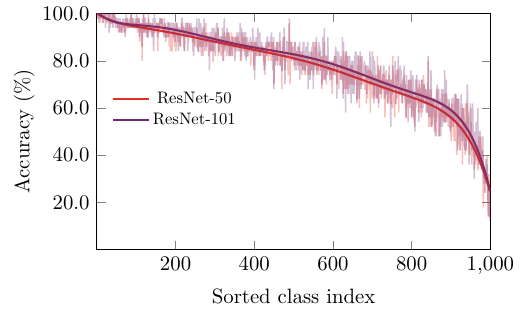}  \label{fig:imagnet_knn}}
    \subfloat[ETF classifier on ImageNet]{ \includegraphics[width=0.22\textwidth]{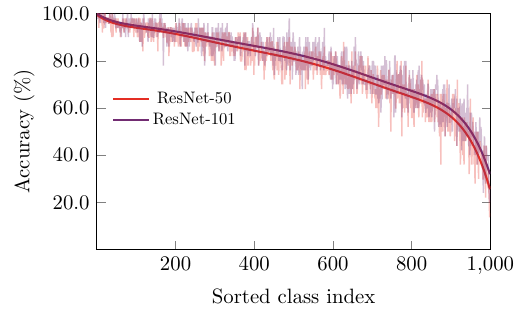}  \label{fig:imagenet_etf}}
    \vspace{-.5em}
    \caption{
        \textbf{Analysis of fairness on bias from representation and classifier.}
        (a) The $\ell_2$-norm of classifier weights on ImageNet;
        (b) Class mean angles defined in Eq.~\eqref{eq:cma};
        (c) Unfairness exists with ResNet-50/101 and $k$-NN algorithm;
        (d) Unfairness exists with ResNet-50/101 and the ETF classifier;
        Classes are sorted by the performance of ResNet-50 on ImageNet.
    }
\label{fig:fairness_analysis_imagenet}
\end{figure}

Interestingly, although the models suffering from unfairness display extremely imbalanced accuracy varying from 100\% to 16\%, we notice that 
the $\ell_2$-norm of classifier weights is already balanced for models trained on ImageNet shown in \Cref{fig:imagenet_weightnorm}, implying that the $\ell_2$-norm of classifier weights is not the core factor for unfairness. 

\begin{definition}
\label{thm:etf}
\normalfont{(Simplex Equiangular Tight Frame).} A collection of vectors $\mathbf{m}_{i} \in \mathbb{R}^d, i=1,2,...,K, d \leq K-1$, is said to be a simplex equiangular tight frame if:
\begin{equation}
    \mathbf{M} = \sqrt{\frac{K}{K-1}} \mathbf{U} \left(\mathbf{I}_{K}- \frac{1}{K} \mathbf{1}_{K} \mathbf{1}^{T}_{K}\right),
\end{equation} 
where $\mathbf{M}=[\mathbf{m}_{1},...,\mathbf{m}_{K}]  \in \mathbb{R}^{d \times K}$ allows a rotation and satisfies $\mathbf{U}^{T}\mathbf{U}=\mathbf{I}_{K}$, $\mathbf{I}_{K}$ is the identity matrix, and $\mathbf{1}_{K}$ is an all-ones vector.

All vectors in a simplex ETF have an equal $\ell_2$-norm and the same pair-wise angle, \ie, $\mathbf{m}_{i}^T \mathbf{m}_{j} = -\frac{1}{K-1}$ if i $\neq$ j. 
\end{definition}

\minisection{Maximal Equiangular Separation}.
The simplex equiangular tight frame (ETF) in \Cref{thm:etf} is introduced by Papyan et al.~\cite{papyan2020prevalence}.
Then several works~\cite{zhong2023understanding, yang2022inducing} explore the ETF classifier for long-tailed recognition. ETF classifier enjoys the same pair-wise angle for any two different classes and allows low-frequency classes to have the maximal equiangular separation. 

We examine if bias in unfairness comes from the smaller separation angles between ``Hard" and other classes by conducting experiments on ImageNet. The mean angle for class $i$, denoted by ${\rm CMA}(i)$, is defined as follows:
\begin{eqnarray}
    {\rm CMA}(i) = \cfrac{1}{C} \sum_{j}^{C} \cfrac{\mathbf{w}_{i} \cdot \mathbf{w}_{j}}{||\mathbf{w}_{i}||_{2} \cdot ||\mathbf{w}_{j}||_{2}} \,,
    \label{eq:cma}
\end{eqnarray} 
where $\mathbf{w}_{i}$ is the classifier weight for class $i$, and C is the number of classes.

As shown in \Cref{fig:imagenet_mean_margin}, most of the ``Hard'' classes get smaller CMA values than ``Easy'' ones and thus they even possess larger separation angles with other classes,
which is quite contrary to our common belief: \textit{larger separation angles usually indicate a higher accuracy}.

We hypothesize that the problematic representations lead to the contrary. When samples in two classes look similar or have overlapping scenarios, the model can be confused. During optimization, the cross-entropy loss will enlarge the separation angles between the two classes and meanwhile encourage their representations more distinguishable from each other. However, if there is little effect on the representations, the two classes still suffer from poor accuracy despite large separation angles between them.

\minisection{Problematic Representation in Unfairness}.
We verify the hypothesis with the following empirical studies:
\begin{itemize}
    \item Evaluate the model with $k$-NN algorithm and examine the existence of unfairness.
    \item Train models with a fixed ETF classifier and examine the existence of unfairness.
\end{itemize}
For the $k$-NN algorithm, with $k$ nearest neighbors in training data, the prediction is determined on the basis of a majority vote, \ie, the label most frequently represented around its neighbors.
Following DINO~\cite{caron2021emerging}, $k$ is set to $20$ for evaluation.  
ETF classifier satisfies the maximal equiangular separation principle. We follow the implementation in previous works~\cite{yang2022inducing, zhong2023understanding}.
As shown in \Cref{fig:imagnet_knn} and \Cref{fig:imagenet_etf}, the class performance still exhibits extreme imbalance, confirming that problematic representation is the core issue for unfairness.

\minisection{Discussion}.
This study highlights the fundamental distinctions between unfairness and long-tailed recognition. Biased classifiers can exacerbate performance disparities in long-tailed recognition. Conversely, unfairness in balanced data stems from problematic representation. 
It provides valuable insights indicating that effective representation learning algorithms are needed to explicitly improve feature quality for the recognition of ``Hard'' classes. Building on these findings, we delve into strategies to enhance fairness in \Cref{sec:improving_fairness}.

\subsection{Model Prediction Bias}
\label{sec:prediction_bias}
Models trained on balanced datasets such as ImageNet exhibit problematic representations,  leading to unfairness. However,  the underlying cause of this issue in terms of optimization remains a mystery. In this section, we are prompted to explore this question with the introduced concept of \textit{Model Prediction Bias} in \Cref{thm:prediction_bias}.

\begin{definition}
	\label{thm:prediction_bias}
	\normalfont{(Model Prediction Bias).} For a well-calibrated~\cite{kumar2022calibrated} classification model $\mathcal{M}$, $x_{i} \in X$ is a input sample. Then the prediction bias for $\mathcal{M}$ is calculated on $X$ with the following formulation:
    \begin{eqnarray}
        \mathbf{G}_{n} &=& \sum_{x_{i} \in X} \sigma (\mathcal{M}(x_{i})) \odot (\mathbf{1}-\mathbf{y}_{i}), \label{eq:Gn}\\
        \mathbf{G}_{p} &=& \sum_{x_{i} \in X} \sigma (\mathcal{M}(x_{i})) \odot \mathbf{y}_{i}, \label{eq:Gp}
    \end{eqnarray}
    where $\sigma$ is the softmax activation. $\mathcal{M}(x_{i}) \in \mathbb{R}^{C}$ is the logits output. $\mathbf{y}_{i}$ is the one-hot label. $C$ is the number of classes. Normalization can be applied to $\mathbf{G}_{n}$ or $\mathbf{G}_{p}$.

    $\mathcal{M}$ is unbiased when $X$ is balanced and $\mathbf{G}_{n}$ is uniform. $\mathbf{G}_{p}$ with the normalization indicates the class accuracy for the well-calibrated model $\mathcal{M}$.
\end{definition} 

\minisection{High Accuracy Partially from Prediction Bias in LT}.
With the concept of \textit{Model Prediction Bias}, we calculate the prediction bias vector $\mathbf{G}_{n}$ of the model trained on ImageNet-LT.
As shown in \Cref{fig:weight_norm_lt}, 
the $\ell_2$-norm 
of classifier weights for high-frequency classes is much larger than that for low-frequency ones.
This leads to the model being biased toward high-frequency classes on all the data (``Many'', ``Medium'', and ``Few'') when giving predictions, as demonstrated in Figs~
\ref{fig:imagenetlt_predictionbias_easy},~\ref{fig:imagenetlt_predictionbias_medium},~\ref{fig:imagenetlt_predictionbias_hard} and~\ref{fig:imagenetlt_predition_bias}.
Thus, the model prediction bias results in a much better performance of high-frequency classes than low-frequency classes.
In a word, in long-tailed recognition, the high performance of high-frequency classes benefits from the prediction bias of the trained model to some degree.

\minisection{Low Accuracy from Prediction Bias for Unfairness}.
On the contrary, as shown in \Cref{fig:imagenet_prediction_bias}, the model suffering from unfairness tends to be biased towards ``Hard" classes which are with poor accuracy. In particular, \textit{the harder the class, the higher the prediction bias the model exhibits.}
We further calculate the prediction bias $\mathbf{G}_{n}$ on ``Easy", ``Medium", and ``Hard" subsets data individually.
Illustrated by \Cref{fig:imagenet_prediction_bias_easy}, we observe that the model has little prediction bias with ``Easy" subset data.
On ``Medium" subset data, the slight tendency of prediction bias towards ``Medium" and ``Hard" classes appears.
\Cref{fig:imagenet_prediction_bias_hard} indicates an obvious prediction bias towards ``Hard" classes of the model when evaluating on the ``Hard" subset data,
which implies that the unfairness mainly comes from the confusion among ``Hard" classes and thus results in their pretty low performance. 

\begin{figure}[tb!]
    \subfloat[ImageNet-LT]{ \includegraphics[width=0.22\textwidth]{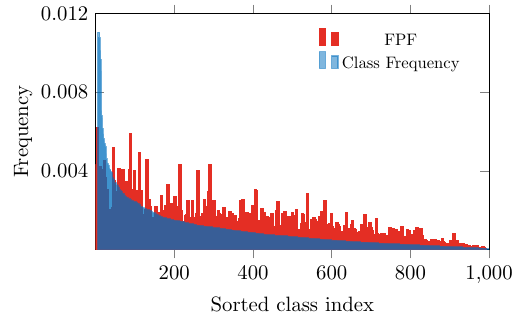}  \label{fig:imagenetlt_fptp}}
    \subfloat[ImageNet]{ \includegraphics[width=0.22\textwidth]{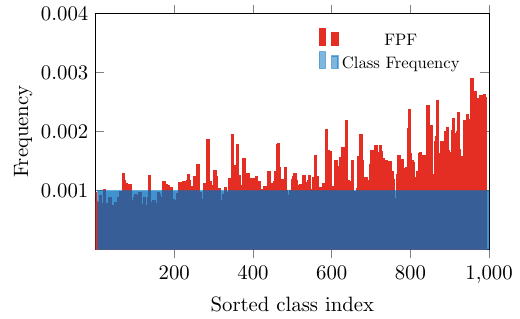}  \label{fig:imagenet_fptp}}
    \vspace{-.7em}
    \caption{
        \textbf{Class frequency vs. False Positive Frequency (FPF).}
        Classes are sorted by the number of samples in the class for ImageNet-LT and the performance of ResNet-50 on ImageNet.
    }
\label{fig:fptp}
\end{figure}

\minisection{High Accuracy or Low Accuracy from Prediction Bias?}
The conclusion appears to be contradictory:
\begin{itemize}
\item  \textit{In long-tail learning, higher prediction biases are found in ``high-frequency'' classes which enjoy high accuracies. In particular, all classes are biased towards high-frequency classes as shown in Figs.~\ref{fig:imagenetlt_predictionbias_easy}, \ref{fig:imagenetlt_predictionbias_medium}, and \ref{fig:imagenetlt_predictionbias_hard}.}
\item \textit{On balanced data, higher prediction biases are observed in ``Hard'' classes which suffer from low accuracies. Specifically, only ``Hard" and ``Medium" classes are biased towards ``Hard" classes in Figs.~\ref{fig:imagenet_prediction_bias_easy},~\ref{fig:imagenet_prediction_bias_medium}, and ~\ref{fig:imagenet_prediction_bias_hard}.}
\end{itemize}

We detail the reasons behind the phenomenon by introducing the following \Cref{thm:gradients_convergence}.

\begin{lemma}
    \label{thm:gradients_convergence}
    \normalfont{(Gradients Convergence Condition).} When the training of model $\mathcal{M}$ converges, the gradients with respect to parameters will be $\mathbf{0}$. Then, the following equation is established.
    \begin{eqnarray}
        \mathbf{G}_{n} + \mathbf{G}_{p} = \mathbf{CF} , \label{eq:gradients_convergence}
    \end{eqnarray}
    where $\mathbf{CF}[i]$ represents the number of samples in class i. 

    \textit{Proof}. With the rule of backpropagation, the gradients with sample $x_{i}$ on $\mathcal{M}(x_{i})$ is what follows,
    \begin{eqnarray}
        \cfrac{\partial \mathcal{L}}{\partial \mathcal{M}(x_{i})} = \sigma(\mathcal{M}(x_{i})) - \mathbf{y}_{i}.
    \end{eqnarray}
    The gradients integrated on the whole training data should be $\mathbf{0}$ for the gradients convergence of training. Then we derive the equation:
    \begin{equation}
        \sum_{x_{i} \!\in\! X} \sigma(\mathcal{M}(x_{i})) -\mathbf{y}_{i} \!=\! \mathbf{0} . \label{eq:gradient}\\
    \end{equation}
    Expand Eq.~\eqref{eq:gradient} with Eq.~\eqref{eq:Gn}, and Eq.~\eqref{eq:Gp},
    \begin{align} 
       \sum_{x_{i} \!\in\! X} \sigma (\mathcal{M}(x_{i})) \!\odot\! (\mathbf{1} \!-\! \mathbf{y}_{i}) \!+\!
        \sum_{x_{i} \!\in\! X} \sigma (\mathcal{M}(x_{i})) \!\odot\! \mathbf{y}_{i} \!=\! \sum_{x_{i} \!\in\! X} \mathbf{y}_{i}.
    \label{eq:lemma_proof}
    \end{align}
    Eq.~\eqref{eq:lemma_proof} is equal to Eq.~\eqref{eq:gradients_convergence}, thus completing the proof.
\end{lemma}

In Lemma~\ref{thm:gradients_convergence}, we theoretically prove the relations among the prediction bias $\mathbf{G}_{n}$, the class performance $\mathbf{G}_{p}$, and the class frequency $\mathbf{CF}$.
As demonstrated by Eq.~\eqref{eq:gradients_convergence}, for long-tailed recognition, the large number of samples ($\mathbf{CF}$) still allows their good performance ($\mathbf{G}_{p}$) for high-frequency classes,
although the model shows higher prediction bias ($\mathbf{G}_{n}$) towards them.
Meanwhile, in terms of the unfairness on balanced data, the higher prediction bias ($\mathbf{G}_{n}$) toward ``Hard'' classes necessarily results in their low accuracy ($\mathbf{G}_{p}$) because the class frequency is uniform ($\mathbf{CF}$).

Intuitively, with \textit{Model Prediction Bias} in \Cref{thm:prediction_bias}, $\mathbf{G}_{n}$ can be regarded as the false positive frequency (FPF) and the class frequency $\mathbf{CF}$ is our expected true positive frequency (TPF).
During training, we improve the score of the ground-truth label and decrease the scores for other classes with cross-entropy loss,
which means we minimize the FPF and expect the TPF to match the class frequency ($\mathbf{CF}$). 
To achieve better accuracy for one specific class, the training should be dominated by true positives (TPs) instead of false positives (FPs) learning.
Otherwise, FPs will overwhelm the learning of TPs, leading to hard optimization and resulting in low performance. 

As shown in \Cref{fig:imagenetlt_fptp}, in long-tailed recognition, the model displays higher $\mathbf{G}_{n}$ towards high-frequency classes. However, TPs still dominate the training due to their high $\mathbf{CF}$.
On the contrary, illustrated by \Cref{fig:imagenet_fptp}, ImageNet enjoys a uniform $\mathbf{CF}$ and the high $\mathbf{G}_{n}$ has overwhelmed their TPs learning for ``Hard" classes, resulting in their hard optimization and poor accuracy.

\minisection{The Unfairness Can Stem from Data Bias}.
We have analyzed the reason for the unfairness phenomenon in terms of training optimization. 
From the perspective of input data, bias in the datasets can lead to unfairness.

Class co-occurrence happens in the ImageNet~\cite{beyer2020we}. For example, ``keyboard" and ``monitor" frequently appear in conjunction. Thus, single-label image annotations in ImageNet can lead to confusion and hard optimization. 
The ImageNet validation set is relabeled as ImageNet-ReaL where multiple labels can be assigned per image. 
Evaluating on ImageNet-ReaL, the accuracy of the worst class improves from 16\% to 45\% but is still far behind the best class 100\%, demonstrating the existence of other potential factors leading to the imbalance.

Moreover, Some class objects occur in much more complex scenes than others. 
Example classes in ``Easy" and ``Hard" subsets on ImageNet are shown in \Cref{fig:imagenet_easy,fig:imagenet_hard} of the supplementary file. 
We calculate ``the $\ell_2$-norm of the class mean feature variance''
with a well-trained ResNet-50 model on ImageNet.
As shown in \Cref{fig:l2_variance_r50_imagenet}, the feature distribution of ``Hard'' classes has a higher variance than that of ``Easy'' classes.
It demonstrates that ``Hard'' classes cover much more complex scenarios which can have some overlaps with other classes.
In this case, without enough training data for the ``Hard'' classes,
the model can overfit some scenario background and lead to prediction confusion with other classes that share similar scenarios.

\subsection{Discussion}

As illustrated in \Cref{sec:fairness_exists}, unfairness is a general problem in image classification models. In this section, we both empirically and theoretically uncover the mystery behind the unfairness from the perspective of gradient optimization during training. Moreover, regarding dataset bias, we observe that features of ``Hard" classes usually have much higher variances than ``Easy" classes as shown by \Cref{fig:l2_variance_r50_imagenet}, implying that data for ``Hard" classes enjoys larger diversity and possesses lower sample density. From this point of view, the unfairness can be considered as a general imbalance problem, \ie, \textit{data diversity imbalance}. Besides, although we usually assume classes are independent, the relations among classes (\eg, class co-occurrence) will make the \textit{data diversity imbalance} heavier. To our knowledge, there are few efforts to explicitly solve the \textit{data diversity imbalance}. We expect our study to raise more attention from the community to the problem.

\begin{figure}[t]
    \subfloat[ImageNet]{ \includegraphics[width=0.22\textwidth]{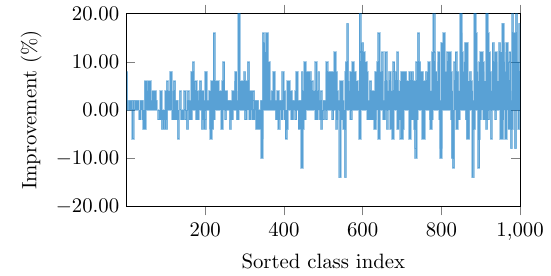}  \label{fig:imagenet_all_delta}}
    \subfloat[CIFAR-100]{ \includegraphics[width=0.22\textwidth]{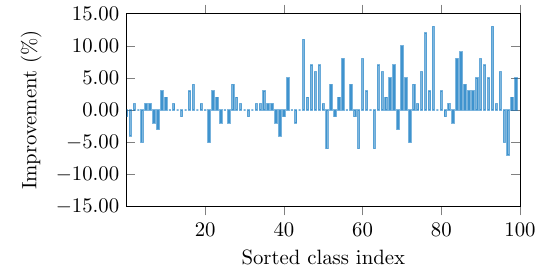}  \label{fig:cifar100_cutmix_delta}}
    \vspace{-.7em}
    \caption{
        \textbf{Improvements of per class with data augmentations on ImageNet and CIFAR-100.}
        (a) Trained with Autoaug, Mixup, and CutMix on ImageNet.
        (b) Trained with CutMix on CIFAR-100.
        ViT-B backbone is used on ImageNet while WideResNet-34-10 is used on CIFAR-100.
    }
\label{fig:aug_per_class}
\end{figure}

\section{Improving Fairness}
\label{sec:improving_fairness}
In this section, we investigate how regular training techniques, like data augmentation and representation learning, affect fairness in image classification. Previous work including Mixup~\cite{zhang2017mixup}, CutMix~\cite{yun2019cutmix}, AutoAug~\cite{cubuk2018autoaugment}, RandAug~\cite{cubuk2020randaugment}, and self-supervised pertaining~\cite{he2022masked} boosts overall performance. However, fairness is rarely discussed and explored. We bridge the gap in this work.

\subsection{Data Augmentation}

\minisection{Mixup and CutMix}.
Mixup~\cite{zhang2017mixup} combines image-label pairs ($x_{i}$, $y_{i}$) and ($x_{j}$,$y_{j}$) with a uniform sampled $\lambda$, 
\begin{eqnarray}
    x_{mix} &=& \lambda * x_{i} + (1-\lambda) * x_{j}, \\
    y_{mix} &=& \lambda * y_{i} + (1-\lambda) * y_{j}, 
\end{eqnarray}
where $\lambda \in [0, 1]$.

CutMix~\cite{yun2019cutmix} extends Mixup and randomly exchanges an image region in $x_{i}$ with another one in image $x_{j}$. Then the mixed label is determined by mixing $y_{i}$ and $y_{j}$ with $\lambda$ which depends on the area of $x_{i}$ region in the mixed image.

\minisection{AutoAug and RandAug.}.
AutoAug~\cite{cubuk2018autoaugment} is one data augmentation strategy that is automatically learned from the data.  
It is learned from a pre-defined search space including various image process operations, like translation, rotation, or shearing. 
RandAug~\cite{cubuk2020randaugment} selects image transformations in a well-designed search space with a uniform probability.

\minisection{Training and Evaluation}.
We use the same training pipeline as described in \Cref{sec:fairness_exists}. For evaluation, we calculate the improvements for each class between models with and without using data augmentation strategies.
We also calculate the improvements of the ``Head", ``Medium", and ``Easy" subsets separately. WideResNet-34-10 is used on CIFAR-100 while ViT-B is adopted for ImageNet. 

\minisection{Results and Analysis}.
\Cref{tab:aug_fairness} shows the improvements on the ``Hard", ``Medium", and ``Easy" subsets on ImageNet and CIFAR-100 datasets with data augmentations. With the experimental results, the consistent observation is that improvements of ``Hard" classes are much larger than that of ``Easy" classes.  
We also plot per-class improvement on CIFAR-100 and ImageNet. As shown in \Cref{fig:aug_per_class}, most of the ``Hard" classes usually enjoy higher performance improvement than ``Easy" classes after adopting data augmentation.  
It indicates that data augmentation strategies can improve the performance fairness of image classification models.
Moreover, data augmentation strategies can be complementary to each other. Combining all of these techniques, the ``Hard" classes obtain more performance gains than applying only a single one.

\subsection{Representation Learning}
\minisection{Contrastive Learning}.
We examine the effects of contrastive learning on fairness in two cases: 1) with the pre-trained weights by MoCo~\cite{he2020momentum} as initialization; 2) using the trained model by GPaCo~\cite{10130611}. For fair comparisons, we use the same training setting in GPaCo~\cite{10130611} for baseline models. ResNet-50 backbone is used in experiments.

\minisection{Masked Modeling}.
Besides contrastive learning, we empirically study how the masked modeling pre-training methods ~\cite{bao2021beit, he2022masked, xie2022simmim} affect fairness in image recognition. We follow the same training pipeline as MAE~\cite{he2022masked} and the ViT-B backbone is adopted in experiments. We also demonstrate that fairness can be boosted by combining contrastive learning and masked modeling pretraining.

\begin{table}[t]
	\centering
	\caption{\textbf{Fairness improvements on ImageNet with contrastive learning and masked modeling}. ``*" represents the model is trained from scratch in 300 epochs.}
	\label{tab:mae_con_fairness} 
    \vspace{-.7em}
    \tablestylesmaller{6.5pt}{1}
		\begin{tabular}{lcccc}
			\toprule
			Method & Easy & Medium & Hard & All      \\
			\midrule
            \multicolumn{5}{c}{Contrastive Learning (ResNet-50)} \\ 
            \midrule
            w/o pre-training        &93.4 &82.0 &62.0 &79.1 \\
            w/ pre-training         &93.4(\textbf{+0.0}) &82.1(\textbf{+0.1}) &62.2(\textbf{+0.2}) &79.2 \\
            GPaCo~\cite{10130611}   &93.8(\textbf{+0.4}) &82.7(\textbf{+0.7}) &62.6(\textbf{+0.6}) &79.7 \\
            \midrule
            \multicolumn{5}{c}{Masked Modeling (ViT-B)} \\ 
            \midrule
            w/o pre-training       &92.2 &79.6 &59.6 &77.1 \\
			w/o pre-training*      &94.3 &84.3 &64.8 &81.1 \\
            w/ pertaining          &95.4(\textbf{+1.1}) &86.6(\textbf{+2.3}) &68.9(\textbf{+4.1}) &83.6 \\
            \midrule
            \multicolumn{5}{c}{Contrastive Learning and Masked Modeling (ViT-B)} \\ 
            \midrule
            GPaCo~\cite{10130611}  &95.5(\textbf{+1.2}) &87.2(\textbf{+2.9}) &69.1(\textbf{+4.3}) &84.0 \\
			\bottomrule
		\end{tabular}
\end{table}

\minisection{Results}.
We list the results in \Cref{tab:mae_con_fairness}. Illustrated by the empirical study, 
both contrastive learning and masked pertaining can achieve better performance gains on ``Hard" classes than ``Easy" classes. Further, we also observe that the masked pertaining~\cite{he2022masked} is more conducive to fairness than contrastive learning. Nevertheless, the two techniques are complementary, and better fairness can be achieved by making them work together.

\subsection{Re-weighting and Other Methods}
Re-weighting plays an important role in long-tailed recognition~\cite{cui2019class, menon2020long}. It assigns larger weights for low-frequency classes and smaller weights for high-frequency ones and expects the low-frequency classes to achieve better performance. We transfer the re-weighting method to address the fairness issue. With a validation set, we calculate the effective frequency (EF) for each class as follows. For class c,
\begin{equation}
    EF(c) = \frac{1}{|X_{c}|} \sum_{x_{i} \in X_{c}} \sigma(\mathcal{M}(x_{i}))[c],
\end{equation}
where $X_{c}$ is a subset of the validation set $X$. Its samples belong to class c. $\sigma$ is the softmax activation.

With the derived effective frequency (EF), we conduct experiments on ImageNet using~\cite{cui2019class, menon2020long}. Besides, we investigate the debiased method LfF~\cite{nam2020learning} and worst-group performance optimization method JTT~\cite{liu2021just} on CIFAR-100.

\noindent{\bf Results.}
As shown in \Cref{tab:reweighting_fairness}, re-weighting methods can improve the accuracy of ``Hard" classes at the cost of some performance degradation on ``Easy" and ``Medium" classes. However, it fails to promote overall accuracy and even achieves inferior performance than the baseline, which is different from data augmentations and representation learning techniques. Moreover, we observe that methods for worst-group performance or classifier debiasing cannot handle complex data distributions, like CIFAR-100.

\begin{table}[t]
	\centering
	\caption{\textbf{Fairness improvements on ImageNet and CIFAR-100 with data augmentation strategies}.}
	\label{tab:aug_fairness} 
    \vspace{-.7em}
    \tablestylesmaller{6pt}{1}
		\begin{tabular}{lcccc}
			\toprule
			Method & Easy & Medium & Hard & All      \\
			\midrule
            \multicolumn{5}{c}{ImageNet} \\ 
            \midrule
            ViT-B      &94.0 &84.1 &64.2  &80.8 \\
			\midrule
            +AutoAug   &94.9(\textbf{+0.9}) &85.9(\textbf{+1.8}) &66.2(\textbf{+2.0}) &82.4 \\
            +Mixup     &94.5(\textbf{+0.5}) &85.3(\textbf{+1.2}) &66.4(\textbf{+2.2}) &82.0 \\
            +CutMix    &94.4(\textbf{+0.4}) &84.6(\textbf{+0.5}) &65.3(\textbf{+1.1}) &81.4 \\
            +All       &95.4(\textbf{+1.4}) &86.6(\textbf{+2.5}) &68.9(\textbf{+4.7}) &83.6 \\
            \midrule
		  \multicolumn{5}{c}{CIFAR-100} \\
            \midrule
            WideResNet-34-10 &92.2 &83.2 &70.1 &81.7 \\
            \midrule
            +AutoAug &91.7(\textbf{-0.5}) &83.5(\textbf{+0.3}) &72.7(\textbf{+2.6}) &82.5 \\
            +Mixup   &91.3(\textbf{-0.9}) &83.5(\textbf{+0.3}) &71.5(\textbf{+1.4}) &82.0 \\
            +CutMix  &92.2(\textbf{+0.0}) &85.0(\textbf{+1.8}) &73.9(\textbf{+3.8}) &83.6 \\
            +Cutout  &92.2(\textbf{+0.0}) &84.2(\textbf{+1.0}) &72.5(\textbf{+2.4}) &82.9 \\
			\bottomrule
		\end{tabular}
\end{table}

\begin{table}[t]
	\centering
	\caption{\textbf{Re-weighting and other methods on fairness.}}
	\label{tab:reweighting_fairness} 
    \vspace{-.7em}
    \tablestylesmaller{4pt}{1}
		\begin{tabular}{lcccc}
			\toprule
			Method & Easy & Medium & Hard & All      \\
                \midrule
                \multicolumn{5}{c}{ImageNet} \\
			\midrule
             Baseline (ResNet-50)  &93.1 &81.1 &59.4  &77.8  \\
             Re-weighting~\cite{menon2020long}  &91.7(\textbf{-1.4})  &79.8(\textbf{-1.3}) &61.4(\textbf{+2.0}) &77.6 \\
             Re-weighting~\cite{cui2019class}   &91.6(\textbf{-1.5})  &79.6(\textbf{-1.5}) &61.3(\textbf{+1.9}) &77.5 \\
             \midrule
             \multicolumn{5}{c}{CIFAR-100} \\
             \midrule
              Baseline (WideResNet34)  &92.2 &83.2 &70.1  &81.7 \\
             \midrule
             Re-weighting~\cite{menon2020long}  &91.5(\textbf{-0.7})  &83.2(\textbf{+0.0}) &70.4(\textbf{+0.3}) &81.6 \\
             LfF~\cite{nam2020learning}         &90.4(\textbf{-1.8}) &81.5(\textbf{-1.7}) &67.7(\textbf{-2.4}) &79.7 \\
             JtT~\cite{liu2021just}             &91.3(\textbf{-0.9}) &82.4(\textbf{-0.8}) &69.6(\textbf{-0.5}) &81.0 \\
		\bottomrule
		\end{tabular}
\end{table}

\section{Conclusion and Future Work}
\vspace{-0.05in}
In this paper, we investigate the image recognition fairness issue, \ie, extreme class performance disparity on balanced data. This unfairness phenomenon widely exists in image classification models.
We identify several properties of the fairness issue by comparing it with long-tailed recognition:
(1) The unfairness comes from problematic representation instead of classifier bias; 
(2) The harder the class, the more other classes can be confused with it, resulting in their low performance. 
Finally, we verify that data augmentation and representation learning tricks can improve fairness with better overall performance.
Re-weighting methods can achieve better fairness by a trade-off between easy and hard classes, without promoting overall accuracy.

With our empirical study, representation learning is promising for improving fairness.
However, it hasn't been explored to explicitly promote fairness learning.
Designing proper data augmentation strategies or self-supervised pre-training techniques specifically for fairness is an interesting direction.
We hope the study will attract more researchers to solve this problem.


{
    \small
    \bibliographystyle{ieeenat_fullname}
    \bibliography{main}
}

\newpage

\onecolumn
\appendix
\begin{center}
	\Large \textbf{Classes Are Not Equal: An Empirical Study on Image Recognition Fairness}
	\Large \\ \textbf{Supplementary Material}
\end{center}

\vspace{+1in}
\section{Class Examples in ImageNet}

\begin{figure*}[h]
    \subfloat[Ostrich]{ \includegraphics[width=0.16\textwidth,height=2.5cm]{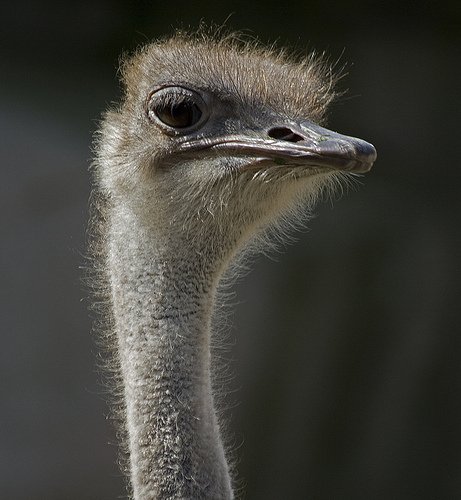}      \label{fig:easy_ostrich1}}
    \subfloat[Ostrich]{ \includegraphics[width=0.16\textwidth,height=2.5cm]{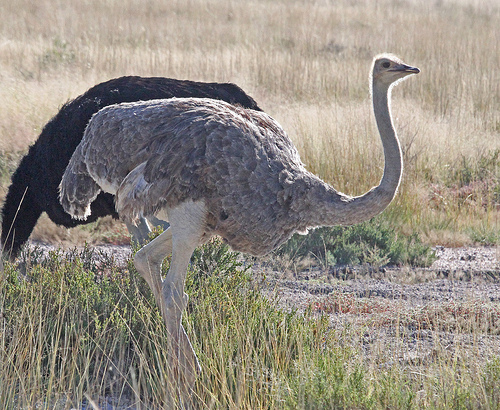}  \label{fig:easy_ostrich2}}
    \subfloat[MacaW]{ \includegraphics[width=0.16\textwidth,height=2.5cm]{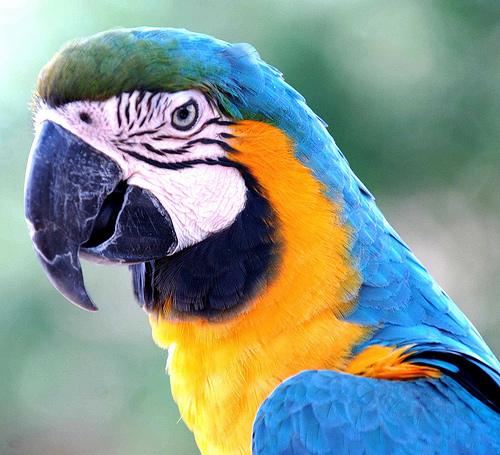} \label{fig:easy_macaw1}}
    \subfloat[MacaW]{ \includegraphics[width=0.16\textwidth,height=2.5cm]{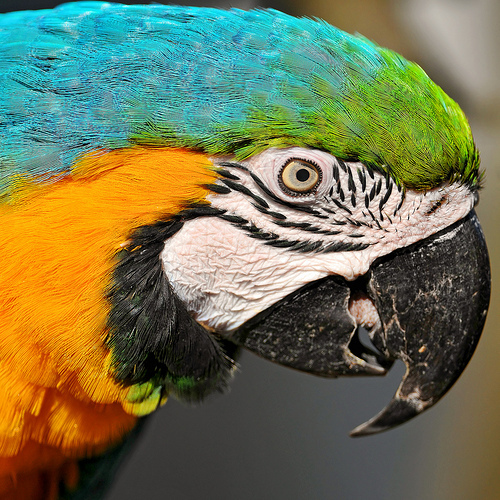}  \label{fig:easy_macaw2}}
    \subfloat[Trilobite]{ \includegraphics[width=0.16\textwidth,height=2.5cm]{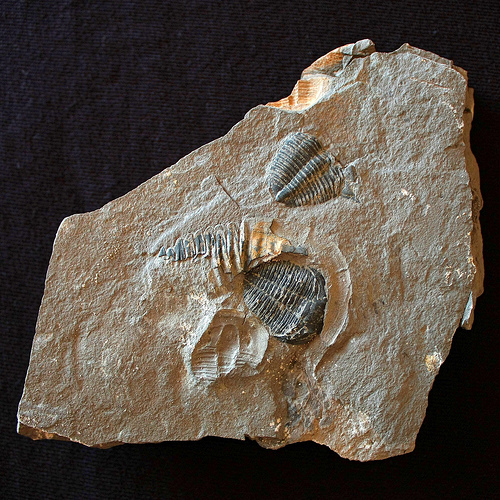} \label{fig:easy_trilobite1}}
    \subfloat[Trilobite]{ \includegraphics[width=0.16\textwidth,height=2.5cm]{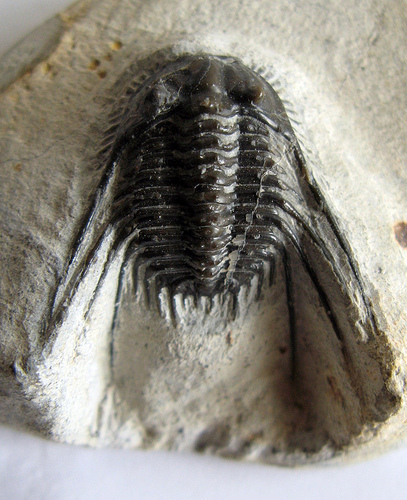} \label{fig:easy_trilobite2}}
    \caption{
        \textbf{``Easy" class examples in ImageNet}
    }
\label{fig:imagenet_easy}
\end{figure*}

\begin{figure*}[h]
    \subfloat[Velvet]{ \includegraphics[width=0.16\textwidth,height=2.5cm]{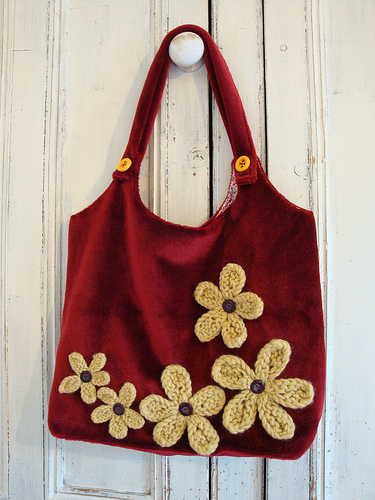}      \label{fig:hard_velvet1}}
    \subfloat[Velvet]{ \includegraphics[width=0.16\textwidth,height=2.5cm]{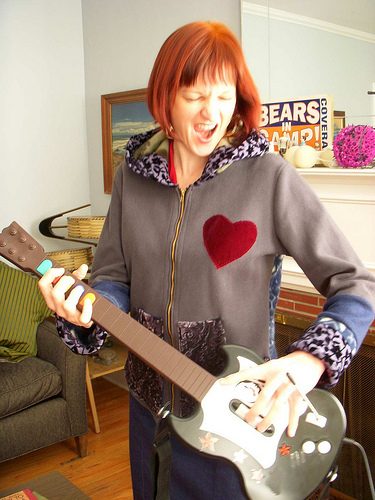}  \label{fig:hard_velvet2}}
    \subfloat[Water jug]{ \includegraphics[width=0.16\textwidth,height=2.5cm]{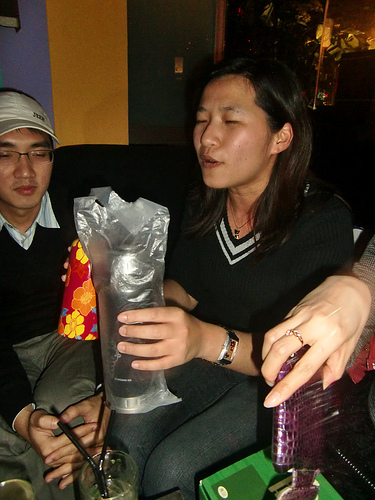} \label{fig:hard_waterjug1}}
    \subfloat[Water jug]{ \includegraphics[width=0.16\textwidth,height=2.5cm]{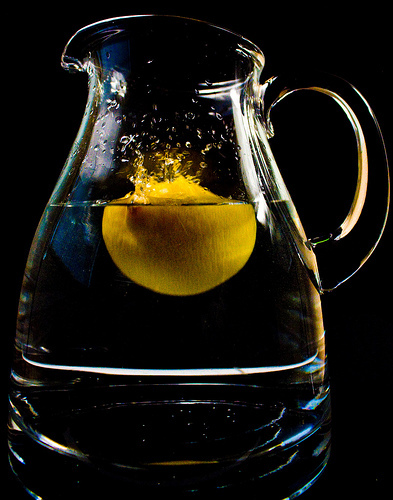}  \label{fig:hard_waterjug2}}
    \subfloat[Screwdriver]{ \includegraphics[width=0.16\textwidth,height=2.5cm]{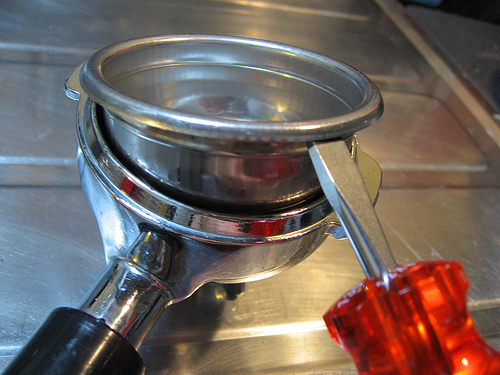} \label{fig:hard_screwdriver1}}
    \subfloat[Screwdriver]{ \includegraphics[width=0.16\textwidth,height=2.5cm]{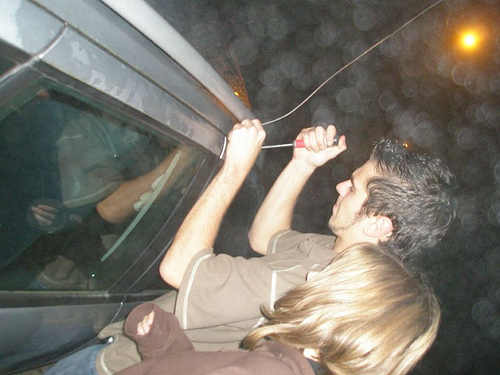} \label{fig:hard_screwdriver2}}
    \caption{
        \textbf{``Hard" class examples in ImageNet}
    }
\label{fig:imagenet_hard}
\vspace{+0.1in}
\end{figure*}

We show class examples in ``Easy" and ``Hard" classes in \Cref{fig:imagenet_easy} and \Cref{fig:imagenet_hard}. ``Easy" classes, like ostrich, macaw, and trilobite, are usually with simple scenarios. However, ``Hard" classes can occur in much more complex scenarios. Take the ``velvet" class as an example. Bags can be made of velvet. Velvet clothes for people or pets also belong to the ``velvet" class. Thus, ``Hard" classes can have overlap scenarios with other classes with a high probability, leading to model confusion and challenging optimization.

\section{Diffusion Classifier}
\paragraph{Oxford-IIIT Pet.}
Oxford-III pet dataset \cite{parkhi2012cats} consists of 37 category pets with roughly 200 images for each class. The images have large variations in scale, pose, and lighting. All images have an associated ground truth annotation of the breed.

\paragraph{Evaluation.}
The stable diffusion model~\cite{rombach2022high} has become one of the most popular foundation models. We examine the fairness of the diffusion classifier \cite{li2023your} on CIFAR-100 and Oxford-IIIT Pet datasets. Checkpoint v2.0 of the stable diffusion model is adopted. With the pre-trained weights, we directly evaluate its zero-shot performance on CIFAR-100 and Oxford-IIIT Pet datasets. All configurations are the same as in diffusion classifier \cite{li2023your}. 

\vspace{+0.1in}
\begin{figure}[h]
\centering
\begin{minipage}{0.45\textwidth}
\centering
\subfloat[CIFAR-100]{\includegraphics[width=0.5\textwidth]{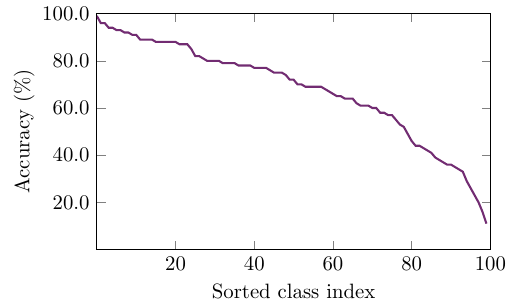} \label{fig:sd_cifar100}}
\subfloat[Oxford-IIIT Pet]{\includegraphics[width=0.5\textwidth]{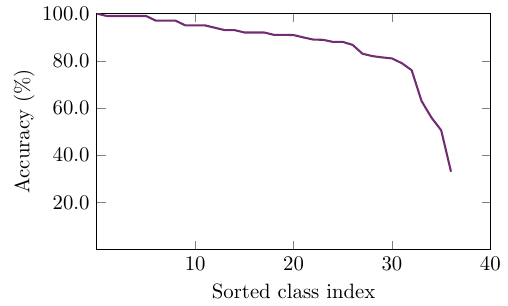} \label{fig:sd_pets}}
\subcaption{Zero-shot class accuracy of the stable diffusion model.}
\end{minipage}
\begin{minipage}{0.45\textwidth}  
\centering
\subcaption{Best and Worst class accuracy of the stable diffusion model.}
\begin{tabular}{ccc}
\toprule
Dataset & Best &Worst\\
\midrule
    CIFAR-100            &99.0 &11.0  \\
    Oxford-IIIT Pet      &100.0 &33.0  \\
\bottomrule
\label{tab:sd}
\end{tabular}
\end{minipage}
\caption{\textbf{Unfairness phenomenon exists in the stable diffusion model.}}
\label{fig:sd}
\vspace{+0.3in}
\end{figure}

\paragraph{Results Analysis.}
The experimental results are listed in \Cref{fig:sd}.
From \Cref{fig:sd_cifar100} and \Cref{fig:sd_pets}, we can see that the class performance exhibits extreme disparity from 99.0\% to 11.0\% on CIFAR-100 and from 100.0\% to 33.0\% on Oxford-IIIT Pet.
Although the stable diffusion model is trained with a huge amount of image-text pair data, it still faces the fairness challenge, demonstrating the prevalence of unfairness in vision-language models.

\vspace{+0.5in}
\section{More Other Datasets}
We demonstrate that performance unfairness is prevalent in image classification. Besides ImageNet, CIFAR, and WIT-400M, other fine-grained benchmarks, including OxfordPets, StandfordCars, Flowers102, Food101, and FGVCAircraft, are also considered in our study.

\paragraph{StanfordCars.}
StanfordCars dataset \cite{krause20133d} contains 16,185 images of 196 classes of cars. The data is split into 8,144 training images and 8,041 testing images, where each class has been split roughly in a 50-50 split. Classes are typically at the level of Make, Model, Year, ex. 2012 Tesla Model S or 2012 BMW M3 coupe.

\paragraph{Flowers102.}
There are 102 flower categories in the Flowers102 dataset \cite{nilsback2008automated}. Each class consists of between 40 and 258 images. The images have large scale, pose, and light variations. In addition, there are categories that have large variations within the category and several very similar categories.

\paragraph{Food101.}
Food101 dataset \cite{bossard14} consists of 101 food categories, with 101,000 images. For each class, 250 manually reviewed test images are provided as well as 750 training images. On purpose, the training images were not cleaned, and thus still contain some amount of noise. This comes mostly in the form of intense colors and sometimes wrong labels.

\paragraph{FGVCAircraft.}
The dataset contains 10,200 images of aircraft, with 100 images for each of 102 different aircraft model variants, most of which are airplanes. It is divided into three equally sized training, validation, and test subsets.

\vspace{+0.2in}
\begin{table*}[h]
    \small
	\centering
	\caption{\textbf{Unfairness phenomenon exists in the popular fine-grained recognition benchmarks}. ``Best" represents the best class accuracy (\%) while ``Worst" denotes the worst class performance.}
	\label{tab:fairness_finegrain} 
    \resizebox{0.9\linewidth}{!}
	{
		\begin{tabular}{lccccccccccc}
		\toprule
		\multirow{2}{*}{Backbone}  &\multicolumn{2}{c}{Oxford-IIIT Pet}  &\multicolumn{2}{c}{StanfordCars} 
             &\multicolumn{2}{c}{Flowers102} &\multicolumn{2}{c}{Food101} & \multicolumn{2}{c}{FGVCAircraft}      \\
             &Best &Worst  &Best &Worst &Best &Worst &Best &Worst &Best &Worst \\
             \midrule
             \multicolumn{11}{c}{Train from scratch} \\
		\midrule 
            ResNet-18 &87.0 &29.2  &97.7 &28.6  &100.0 &4.8 &98.4 &48.8 &97.1 &20.6 \\
            ResNet-34 &95.0 &37.1  &100.0 &31.4 &100.0 &5.6 &99.2 &52.0 &97.0 &21.2 \\
            ResNet-50 &79.0 &16.0  &95.3 &31.0  &100.0 &4.7 &98.4 &55.2 &91.2 &9.0 \\
            \midrule
            \multicolumn{11}{c}{Train with initialization from ImageNet pre-train weights} \\
		\midrule 
            ResNet-18 &98.0 &41.0 &100.0 &44.8 &100.0 &50.0 &100.0 &57.2 &100.0 &33.3 \\
            ResNet-34 &98.0 &43.0 &100.0 &44.8 &100.0 &40.0 &98.4 &51.2  &100.0 &45.4 \\
            ResNet-50 &98.0 &56.0 &100.0 &40.0 &100.0 &50.0 &99.2 &48.4  &100.0 &39.3 \\
		\bottomrule
		\end{tabular}
	}
\end{table*}

\begin{figure}[t]
    \subfloat[ResNet-18]{ \includegraphics[width=0.16\textwidth]{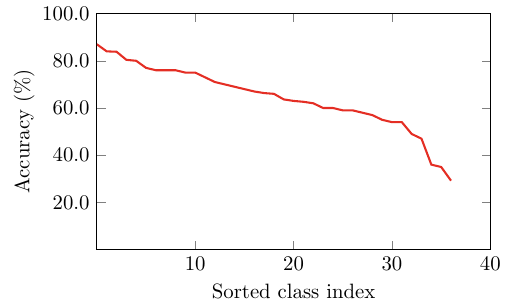}  \label{fig:oxford_nopretrain_r18}}
    \subfloat[ResNet-34]{  \includegraphics[width=0.16\textwidth]{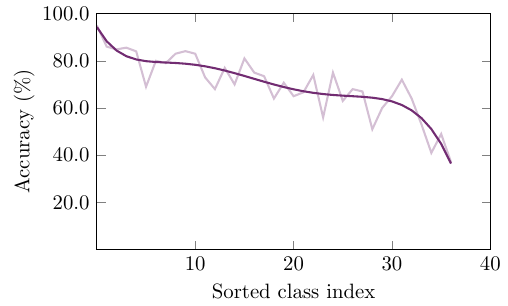}  \label{fig:fairness_nopretrain_r34}}
    \subfloat[ResNet-50]{  \includegraphics[width=0.16\textwidth]{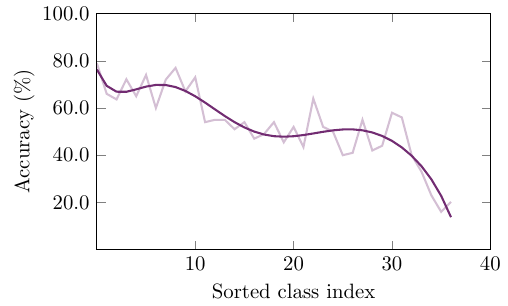}      \label{fig:fairness_nopretrain_r50}}
    \subfloat[ResNet-18]{ \includegraphics[width=0.16\textwidth]{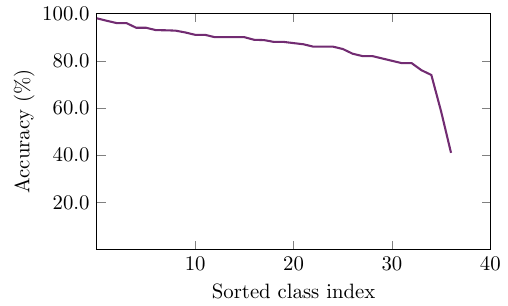}  \label{fig:oxford_pretrain_r18}}
    \subfloat[ResNet-34]{  \includegraphics[width=0.16\textwidth]{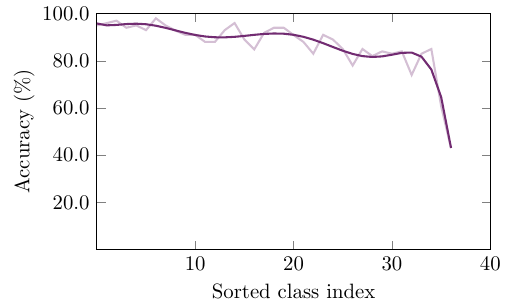}  \label{fig:fairness_pretrain_r34}}
    \subfloat[ResNet-50]{  \includegraphics[width=0.16\textwidth]{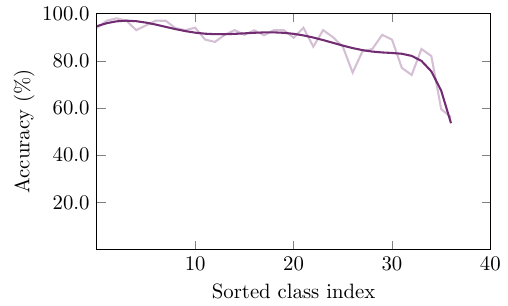}      \label{fig:fairness_pretrain_r50}}
    \caption{
        \textbf{Unfairness on the Oxford-IIIT Pet dataset.}
        (a), (b), and (c) are trained from scratch.
        (d), (e), and (f) are initialized with ImageNet pre-train weights.
    }
\label{fig:fairness_oxfordpets}
\vspace{+0.4in}
\end{figure}

\begin{figure}[t]
    \subfloat[ResNet-18]{ \includegraphics[width=0.16\textwidth]{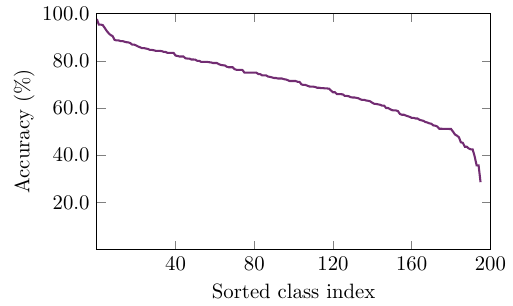}  \label{fig:stanfordcars_nopretrain_r18}}
    \subfloat[ResNet-34]{  \includegraphics[width=0.16\textwidth]{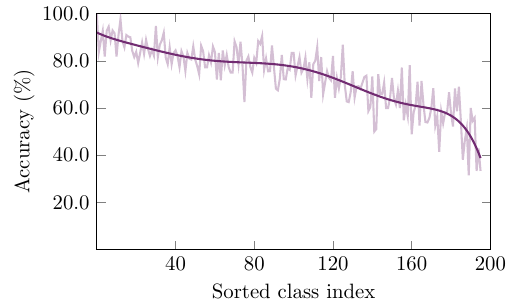}  \label{fig:stanfordcars_nopretrain_r34}}
    \subfloat[ResNet-50]{  \includegraphics[width=0.16\textwidth]{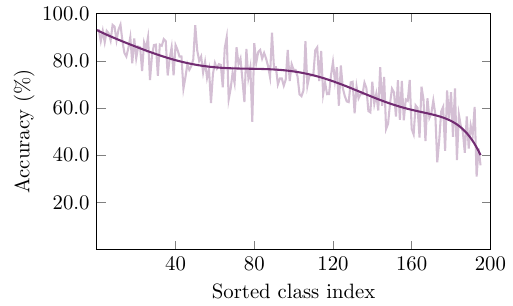}      \label{fig:stanfordcars_nopretrain_r50}}
    \subfloat[ResNet-18]{ \includegraphics[width=0.16\textwidth]{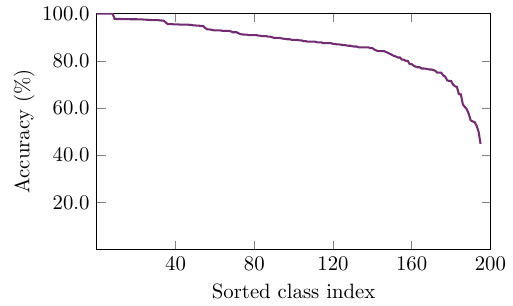}  \label{fig:stanfordcars_pretrain_r18}}
    \subfloat[ResNet-34]{  \includegraphics[width=0.16\textwidth]{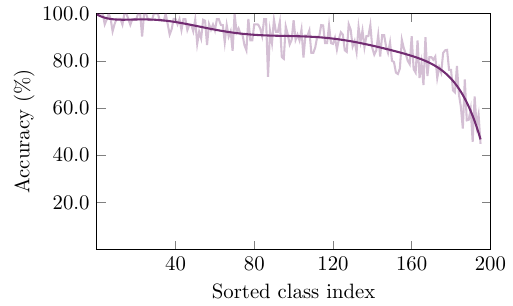}  \label{fig:stanfordcars_pretrain_r34}}
    \subfloat[ResNet-50]{  \includegraphics[width=0.16\textwidth]{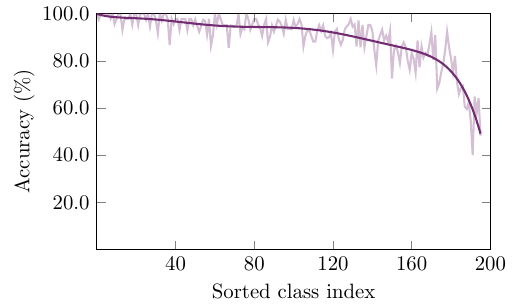}      \label{fig:stanfordcars_pretrain_r50}}
    \caption{
        \textbf{Unfairness on the StanfordCars dataset.}
        (a), (b), and (c) are trained from scratch.
        (d), (e), and (f) are initialized with ImageNet pre-train weights.
    }
\label{fig:fairness_stanfordcars}
\vspace{+0.4in}
\end{figure}

\begin{figure}[t]
    \subfloat[ResNet-18]{ \includegraphics[width=0.16\textwidth]{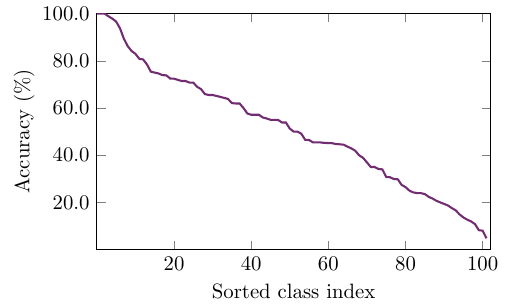}  \label{fig:flowers102_nopretrain_r18}}
    \subfloat[ResNet-34]{  \includegraphics[width=0.16\textwidth]{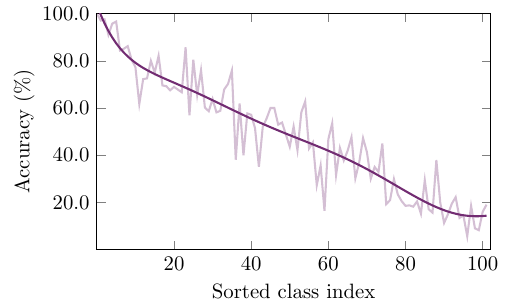}  \label{fig:flowers102_nopretrain_r34}}
    \subfloat[ResNet-50]{  \includegraphics[width=0.16\textwidth]{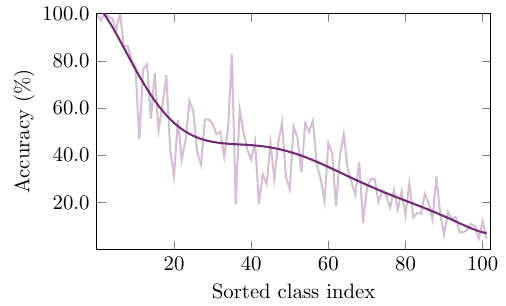}      \label{fig:flowers102_nopretrain_r50}}
    \subfloat[ResNet-18]{ \includegraphics[width=0.16\textwidth]{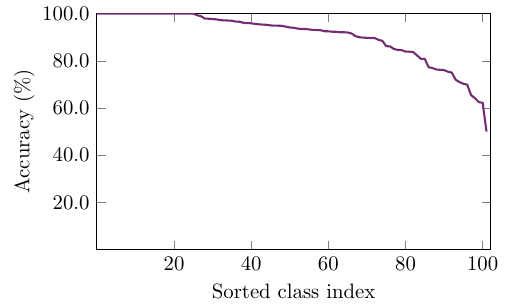}  \label{fig:flowers102_pretrain_r18}}
    \subfloat[ResNet-34]{  \includegraphics[width=0.16\textwidth]{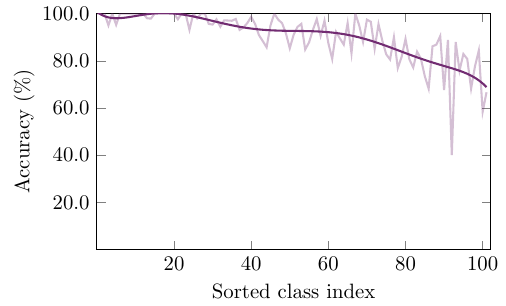}  \label{fig:flowers102_pretrain_r34}}
    \subfloat[ResNet-50]{  \includegraphics[width=0.16\textwidth]{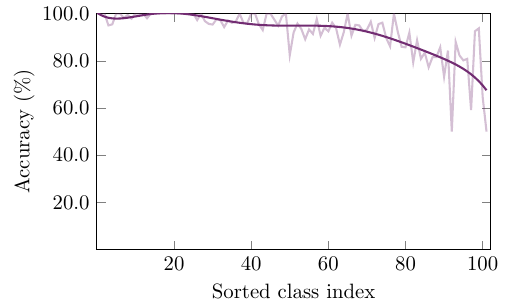}      \label{fig:flowers102_pretrain_r50}}
    \caption{
        \textbf{Unfairness on the Flowers102 dataset.}
        (a), (b), and (c) are trained from scratch.
        (d), (e), and (f) are initialized with ImageNet pre-train weights.
    }
\label{fig:fairness_flowers102}
\vspace{+0.3in}
\end{figure}

\begin{figure}[t]
    \subfloat[ResNet-18]{ \includegraphics[width=0.16\textwidth]{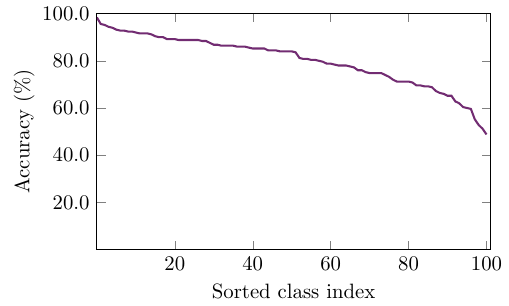}  \label{fig:food101_nopretrain_r18}}
    \subfloat[ResNet-34]{  \includegraphics[width=0.16\textwidth]{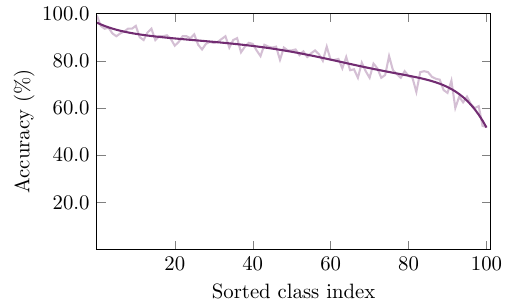}  \label{fig:food101_nopretrain_r34}}
    \subfloat[ResNet-50]{  \includegraphics[width=0.16\textwidth]{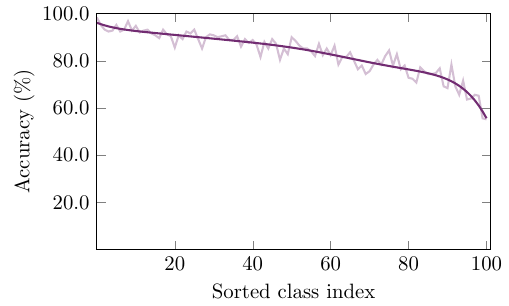}      \label{fig:food101_nopretrain_r50}}
    \subfloat[ResNet-18]{ \includegraphics[width=0.16\textwidth]{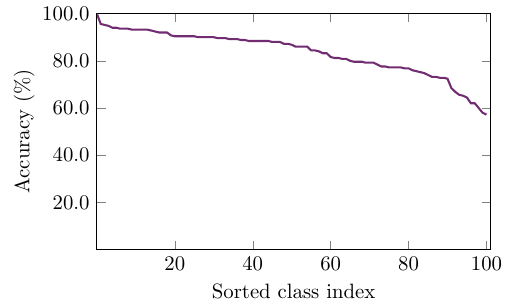}  \label{fig:food101_pretrain_r18}}
    \subfloat[ResNet-34]{  \includegraphics[width=0.16\textwidth]{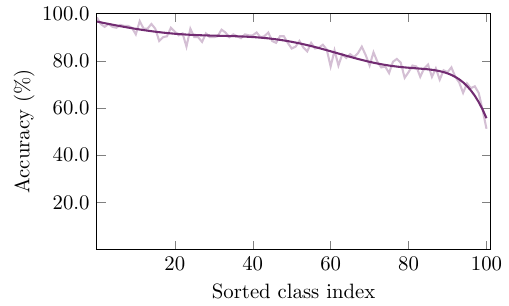}  \label{fig:food101_pretrain_r34}}
    \subfloat[ResNet-50]{  \includegraphics[width=0.16\textwidth]{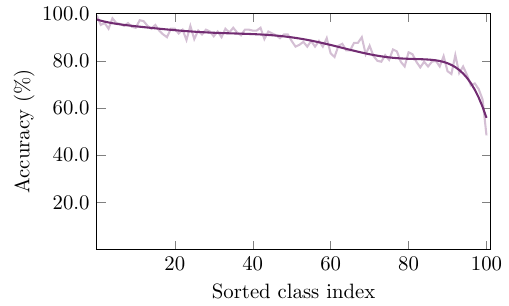}      \label{fig:food101_pretrain_r50}}
    \caption{
        \textbf{Unfairness on the Food101 dataset.}
        (a), (b), and (c) are trained from scratch.
        (d), (e), and (f) are initialized with ImageNet pre-train weights.
    }
\label{fig:fairness_food101}
\vspace{+0.3in}
\end{figure}

\begin{figure}[t]
    \subfloat[ResNet-18]{ \includegraphics[width=0.16\textwidth]{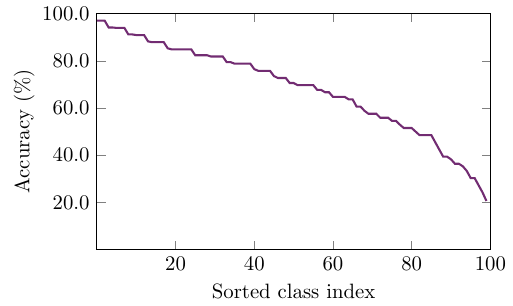}  \label{fig:fgvc_nopretrain_r18}}
    \subfloat[ResNet-34]{  \includegraphics[width=0.16\textwidth]{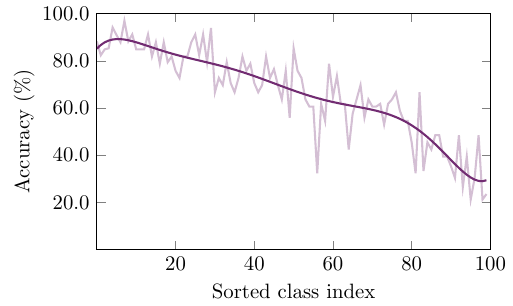}  \label{fig:fgvc_nopretrain_r34}}
    \subfloat[ResNet-50]{  \includegraphics[width=0.16\textwidth]{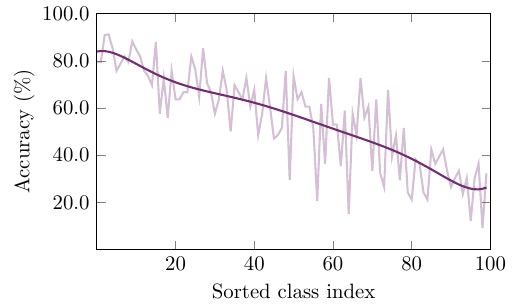}      \label{fig:fgvc_nopretrain_r50}}
    \subfloat[ResNet-18]{ \includegraphics[width=0.16\textwidth]{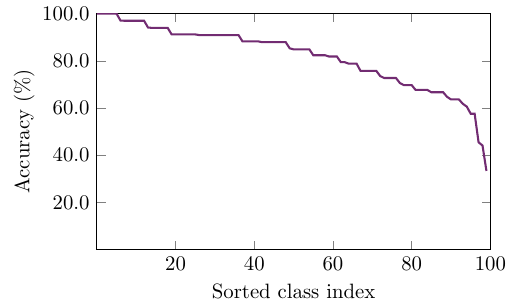}  \label{fig:fgvc_pretrain_r18}}
    \subfloat[ResNet-34]{  \includegraphics[width=0.16\textwidth]{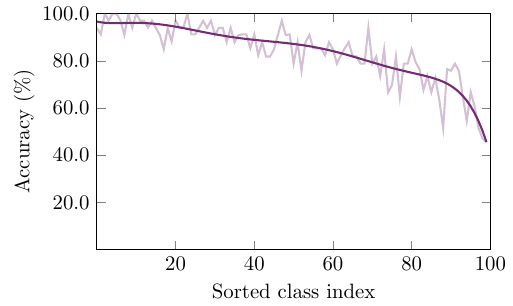}  \label{fig:fgvc_pretrain_r34}}
    \subfloat[ResNet-50]{  \includegraphics[width=0.16\textwidth]{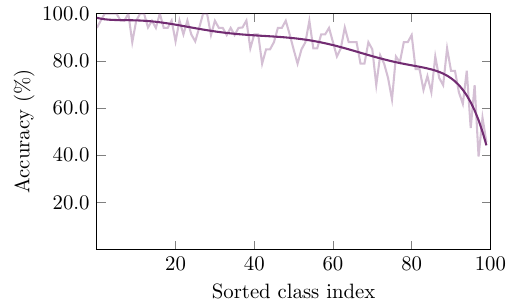}      \label{fig:fgvc_pretrain_r50}}
    \caption{
        \textbf{Unfairness on the FGVCAircraft dataset.}
        (a), (b), and (c) are trained from scratch.
        (d), (e), and (f) are initialized with ImageNet pre-train weights.
    }
\label{fig:fairness_fgvc}
\vspace{+0.5in}
\end{figure}

\newpage
\paragraph{Training and Evaluation.}
We use ResNet-18, ResNet-34, and ResNet-50 as our backbones. Following the training schedule on ImageNet, we use the same pre-process, \ie, randomly crop and resize to $224 \times 224$ and then randomly horizontal flip.
Models are trained in 100 epochs with a cosine learning rate strategy and a batch size of 256 on 8 GPUs. The initial learning rate and the weight decay are set to 0.1 and 1e-4 separately. The SGD optimizer with a momentum of 0.9 is used. 

We evaluate the performance of models trained from scratch and initialized with ImageNet pre-train weights.
ImageNet pre-train weights embed training data information of ImageNet. Thus,
initialization with the pre-train weights can disturb the original training data distribution. 
Considering that, we use a ResNet-18 model trained from scratch as the reference model to sort classes under the case without ImageNet pre-train weight initialization. Otherwise, we use a ResNet-18 model trained with the weight initialization as the reference model to sort classes.
As shown in Figures~\ref{fig:fairness_cifar100_imagenet}, ~\ref{fig:fairness_oxfordpets},~\ref{fig:fairness_stanfordcars},~\ref{fig:fairness_flowers102},~\ref{fig:fairness_food101},~\ref{fig:fairness_fgvc}, various models exhibit similar trends on the same dataset, implying that the unfairness highly depends on training data distribution.

\paragraph{Results Analysis.}
Our results are summarized in Figs~\ref{fig:fairness_oxfordpets},~\ref{fig:fairness_stanfordcars},~\ref{fig:fairness_flowers102},~\ref{fig:fairness_food101},~\ref{fig:fairness_fgvc} and \Cref{tab:fairness_finegrain}. For models training from scratch, the performance unfairness is obvious on Oxford-IIIT pet, StandardCars, Flowers102, and FGVCAircraft datasets. Particularly, there is over 70\% performance disparity between the best class and the worst class on the FGVCAircraft dataset, demonstrating the severe fairness issue.
On models trained with initialization from ImageNet pre-train weights, the worst class performance significantly increases. However, the extreme performance imbalance still exists, specifically on Oxford-IIIT Pet, StanfordCars, and FGVCAircraft datasets.

Without initialization from ImageNet pre-train weights, we observe that the model performance can decrease as the capacity increases. The accuracy of the ResNet-50 model is lower than that of ResNet-18 and ResNet-34 on the Oxford-IIIT Pet dataset. This phenomenon can be caused by limited training data.

\section{Equalized Odds Evaluation}
Following the definition of Equalized Odds (EO), we extend it with a tighter constrain:
\vspace{-0.1in}
\begin{equation}
    \scriptsize
    P(\hat Y=y_{i} | Y=y_{i}, A=y_{i}) = P(\hat Y=y_{j} | Y=y_{j}, A=y_{j}),
    \vspace{-0.1in}
\end{equation}
where $y_{i}$, $y_{j}$ $\in$ {1,2,...,C}. C is the number of classes. $\hat Y$ is the prediction. $Y$ is the true label, and $A$ refers to group membership. Here, we treat classes as groups. We report the maximum False Positive Error Rate (FPR) and False Negative Error Rate (FNR) disparities among $C$ groups in \Cref{tab:eo}.

\begin{table}[h]
	\centering
	\caption{\textbf{EO for fairness on ImageNet.}}
	\label{tab:eo_fairness} 
	{
		\begin{tabular}{lc@{\ \ }c@{\ \ }c}
			\toprule
			  EO metrics      & ResNet-50 & ResNet-101 & ViT-B    \\
                \midrule
                FPR balance  &0.78 &0.74 &0.80 \\
                FNR balance  &0.84 &0.84 &0.80 \\
		      \bottomrule
		\end{tabular}
	}
\label{tab:eo}
\end{table}

\end{document}